%% file: main.tex
\documentclass[10pt]{article}
\usepackage[dvipsnames,table]{xcolor}
\usepackage[accepted]{tmlr}
\usepackage{graphicx}
\usepackage{multirow}
\usepackage{rotating}
\usepackage{booktabs}
\usepackage{amssymb}
\usepackage{subcaption}
\usepackage{ulem}
\usepackage{tikz, pgfmath}
\usepackage{style}
\usepackage{soul}
\usepackage{hyperref}
\usepackage{enumitem}
\usepackage{wrapfig} 
\newcolumntype{?}{!{\color{white}\vrule width 6pt}}

\definecolor{lightergray}{cmyk}{0.00, 0.00, 0.00, 0.14}
\DeclareRobustCommand{\hllightgray}[1]{{\sethlcolor{lightergray}\hl{#1}}}
\DeclareRobustCommand{\hlgray}[1]{{\sethlcolor{lightgray}\hl{#1}}}
\definecolor{lemon}{cmyk}{0., 0., 0.41, 0.01}
\definecolor{vibgreen}{rgb}{0.455, 0.733, 0.290}

\DeclareRobustCommand{\yellowcircle}{\tikz{\draw[fill=yellow, draw=none] (0,0) circle (3pt);}}
\DeclareRobustCommand{\bluecircle}{\tikz{\draw[fill=blue, draw=none] (0,0) circle (3pt);}}
\DeclareRobustCommand{\greencircle}{\tikz{\draw[fill=vibgreen, draw=none] (0,0) circle (3pt);}}
\DeclareRobustCommand{\orangecircle}{\tikz{\draw[fill=orange, draw=none] (0,0) circle (3pt);}}

\definecolor{g1}{HTML}{00994C}
\definecolor{g2}{HTML}{19A35C}
\definecolor{g3}{HTML}{33AD6B}
\definecolor{g4}{HTML}{4DB77A}
\definecolor{g5}{HTML}{66C189}
\definecolor{g6}{HTML}{80CBA0}
\definecolor{g7}{HTML}{99D5B4}
\definecolor{g8}{HTML}{B3DFC8}
\definecolor{g9}{HTML}{CCEADC}
\definecolor{g10}{HTML}{E6F5F0}
\graphicspath{{Figures/}}
\DeclareGraphicsExtensions{.pdf,.jpeg,.png}

\usepackage{amsmath}
\usepackage{bbm}
\usepackage[thinc]{esdiff}
\usepackage{algorithmic}
\usepackage{array}
\usepackage{stfloats}
\usepackage{booktabs,arydshln}
\usepackage{adjustbox}

\makeatletter
\def\adl@drawiv#1#2#3{%
        \hskip.5\tabcolsep
        \xleaders#3{#2.5\@tempdimb #1{1}#2.5\@tempdimb}%
                #2\z@ plus1fil minus1fil\relax
        \hskip.5\tabcolsep}
\newcommand{\cdashlinelr}[1]{%
  \noalign{\vskip\aboverulesep
           \global\let\@dashdrawstore\adl@draw
           \global\let\adl@draw\adl@drawiv}
  \cdashline{#1}
  \noalign{\global\let\adl@draw\@dashdrawstore
           \vskip\belowrulesep}}
\makeatother

\usepackage{url}
\newcommand{\blockcomment}[1]{}
\usepackage[capitalise]{cleveref}
\usepackage[pro]{fontawesome5}
\usepackage{cuted}

\newcommand{\edit}[1]{\textcolor{black}{#1}}
\newcommand{\element}[1]{component}
\newcommand{\module}[1]{module}
\newcommand{\Element}[1]{component}
\newcommand{\Module}[1]{Module}

\title{Deep Sprite-based Image Models: An Analysis}

\author{\name Zeynep Sonat Baltacı \email{sonat.baltaci@enpc.fr} \\
    \addr LIGM, CNRS, Univ Gustave Eiffel, ENPC, Institut Polytechnique de Paris, France
    \AND
    \name Romain Loiseau \email{romain.loiseau@enpc.fr} \\
    \addr LIGM, CNRS, Univ Gustave Eiffel, ENPC, Institut Polytechnique de Paris, France
    \AND
    \name Mathieu Aubry \email{mathieu.aubry@enpc.fr} \\
    \addr LIGM, CNRS, Univ Gustave Eiffel, ENPC, Institut Polytechnique de Paris, France
}

\def\month{06}  
\def\year{2026} 
\def\openreview{\url{https://openreview.net/forum?id=pXuxMLFo9g}} 

\begin{document}
\maketitle

\begin{abstract}
While foundation models drive steady progress in image segmentation and diffusion algorithms compose always more realistic images, the seemingly simple problem of identifying recurrent patterns in a collection of images remains very much open. In this paper, we focus on sprite-based image decomposition models, which have shown some promise for clustering and image decomposition and are appealing because of their high interpretability. These models come in different flavors, need to be tailored to specific datasets, and struggle to scale to images with many objects. 
We dive into the details of their design, identify their core components, and perform an extensive analysis on clustering benchmarks. We leverage this analysis to propose a deep sprite-based image decomposition method that performs on par with state-of-the-art unsupervised \edit{class-aware} image segmentation methods on the standard CLEVR benchmark, scales linearly with the number of objects, identifies explicitly object categories, and fully models images in an easily interpretable way.\footnote{\url{https://github.com/sonatbaltaci/deepsprite}}
\end{abstract}


\input{Sections/1-Introduction}
\input{Sections/2-RelatedWork}
\input{Sections/3-Method}
\input{Sections/4-Analysis}
\input{Sections/5-MultiObject}
\input{Sections/6-Conclusion}

\section*{Acknowledgments}
Z. S. Baltac\i{} and M. Aubry supported by the ANR project VHS ANR-21-CE38-0008, and the ERC project DISCOVER funded by the European Union's Horizon Europe Research and Innovation program under grant agreement No. 101076028. This work was granted access to the HPC resources of IDRIS under the allocation AD011015415R1, AD011015415, and AD011014404 made by GENCI. We would like to thank Ioannis Siglidis for insightful discussions, and Robin Champenois and Ségolène Albouy for their contributions to the codebase.
\bibliographystyle{tmlr}
\bibliography{ref}

\newpage
\appendix
\input{Sections/7-Appendix}
\end{document}

%% file: Sections/1-Introduction.tex
\begin{figure}[ht!]
    \centering
    \begin{subfigure}[]{0.75\columnwidth}
    \centering
        \includegraphics[width=\columnwidth, trim=2cm 10cm 2cm 10cm,
        clip]{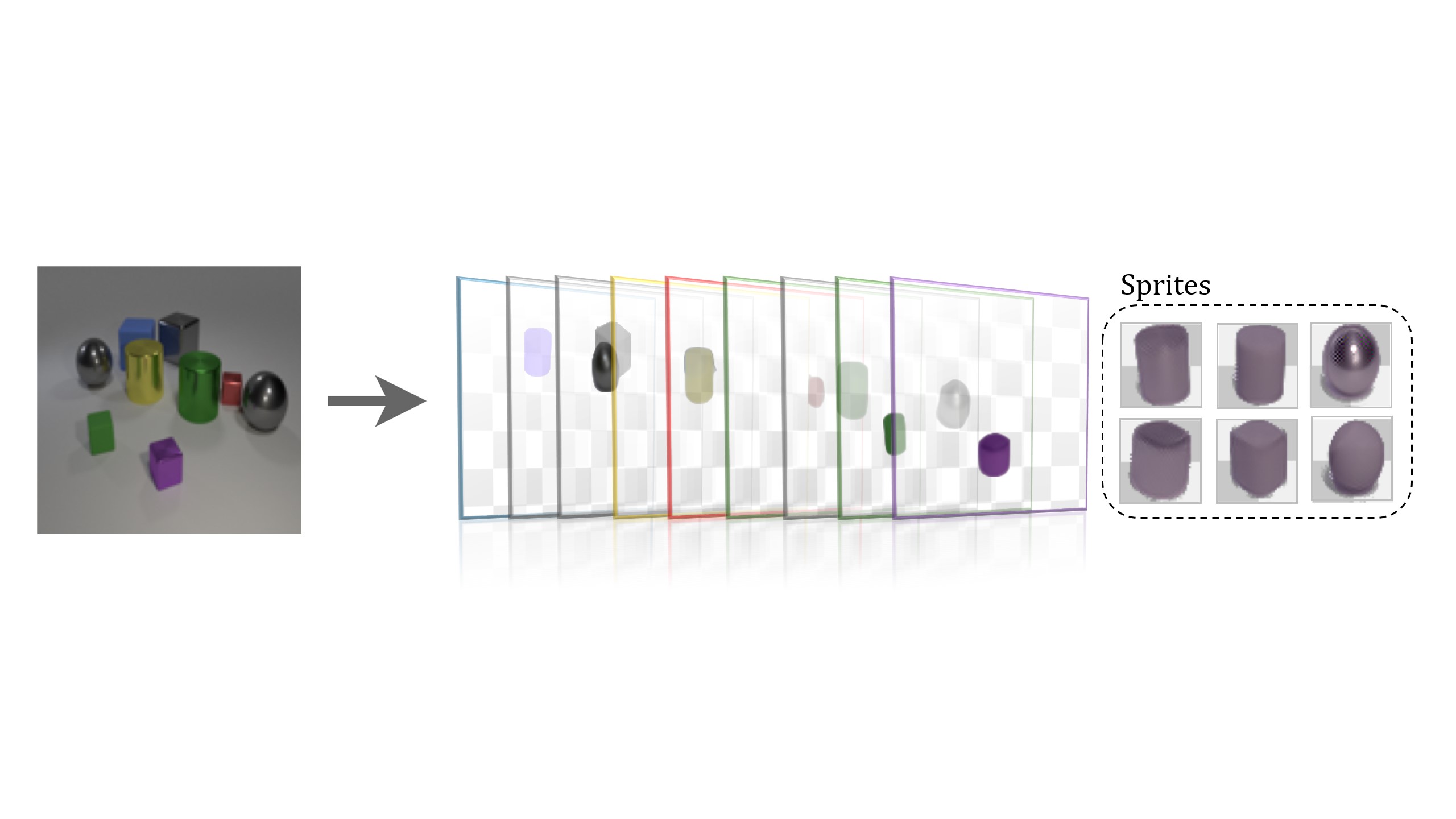}
        \caption{\textbf{Sprite-based approaches.}}
        \label{subfig:teaser-clustering}
    \end{subfigure}
    \vfill \vspace{5pt}
    \begin{subfigure}[]{0.45\columnwidth}
    \centering
        \includegraphics[width=\columnwidth]{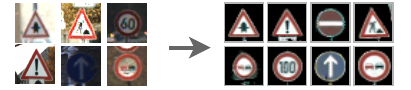}
        \caption{\textbf{Image clustering.}}
        \label{subfig:teaser-gtsrb}
    \end{subfigure}
    \begin{subfigure}[]{0.45\columnwidth}
    \centering
        \includegraphics[width=\columnwidth]{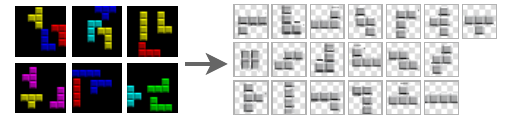}
        \caption{\textbf{Unsupervised object discovery.}}
        \label{subfig:teaser-tetro}
    \end{subfigure}
    \caption{ (a) Sprite-based approaches take a set of images as input and learn jointly a family of sprites and how to decompose each image into a sequence of transformed sprites. They can be applied to  (b) image clustering and (c) unsupervised object discovery.}
    \label{fig:teaser}
\end{figure}
\section{Introduction}
\label{sec:intro}
Identifying recurring patterns in an image collection is a task in which humans excel. It is also critical for many scientific applications, from historical documents to medical image analysis. Although foundation features or models might be attractive tools for approaching this problem, they come with their black-box effects and the biases of their training data. Instead, we advocate for methods that can be directly optimized on the target image collection, offer maximal interpretability, and have limited bias.

In this study, we focus more specifically on sprite-based methods~\citep{visser2019stampnet,monnier2020dticlustering,monnier2021unsupervised,smirnov2021marionette,loiseau2023learnable,siglidis2024ltr}, which are the main type of object-centric approaches to unsupervised object discovery that allow joint categorization and localization~\citep{villa2024unsupervised} (\cref{fig:teaser}). Sprite-based methods offer several other attractive advantages. First, they explicitly model repeated patterns as a finite set of prototypical objects, called sprites. Second, not only do they provide for each analyzed image a layered decomposition, but they also give direct, explicit access to the transformation of the sprites in the image, such as position, scale, and color transformations. Third, their relationship with the standard K-means clustering algorithm~\citep{macqueen1967some,bottou1994convergence} and transformation invariant methods~\citep{frey99em,frey01fltic,frey03tic} is well understood~\citep{monnier2020dticlustering}.
\begin{figure*}[t]
\centering
\includegraphics[width=\textwidth]{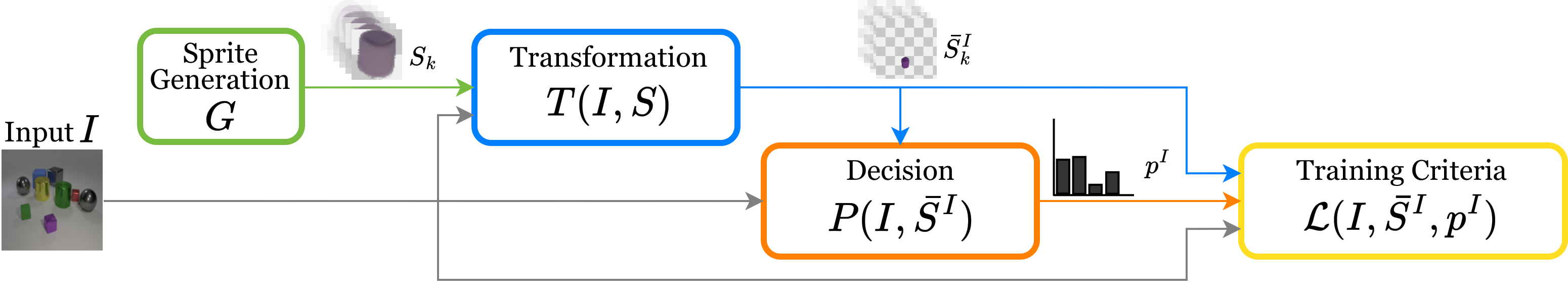}
\caption{\textbf{Overview.} We decompose all sprite-based models in four main components: (1) a \textit{Sprite Generation Module} (\greencircle{}) that outputs $K$ sprites $S_{}$, (2) a \textit{Transformation Module} (\bluecircle{}) that takes as input an image $I$ and the sprites $S$ to predict transformed sprites $\bar{S}^I$, (3) a \textit{Decision Module} (\orangecircle{}) that takes the image $I$ and transformed sprites $\bar{S}^I$ as input and outputs a probability distribution $p^I$ for using the sprites, and (4) a \textit{Training Criteria} (\yellowcircle{}) which consist of a reconstruction loss and potential regularization terms.}
\label{fig:arch}
\end{figure*}
However, sprite-based methods have not been fully explored. In particular, the impact of architectural changes and training methodology on their results is poorly understood and different approaches have been demonstrated on different non-standard datasets. Our goal in this study is to better identify key design choices for sprite-based methods and analyze their effects.

\begin{figure*}[t]
  \centering
\includegraphics[width=\textwidth]{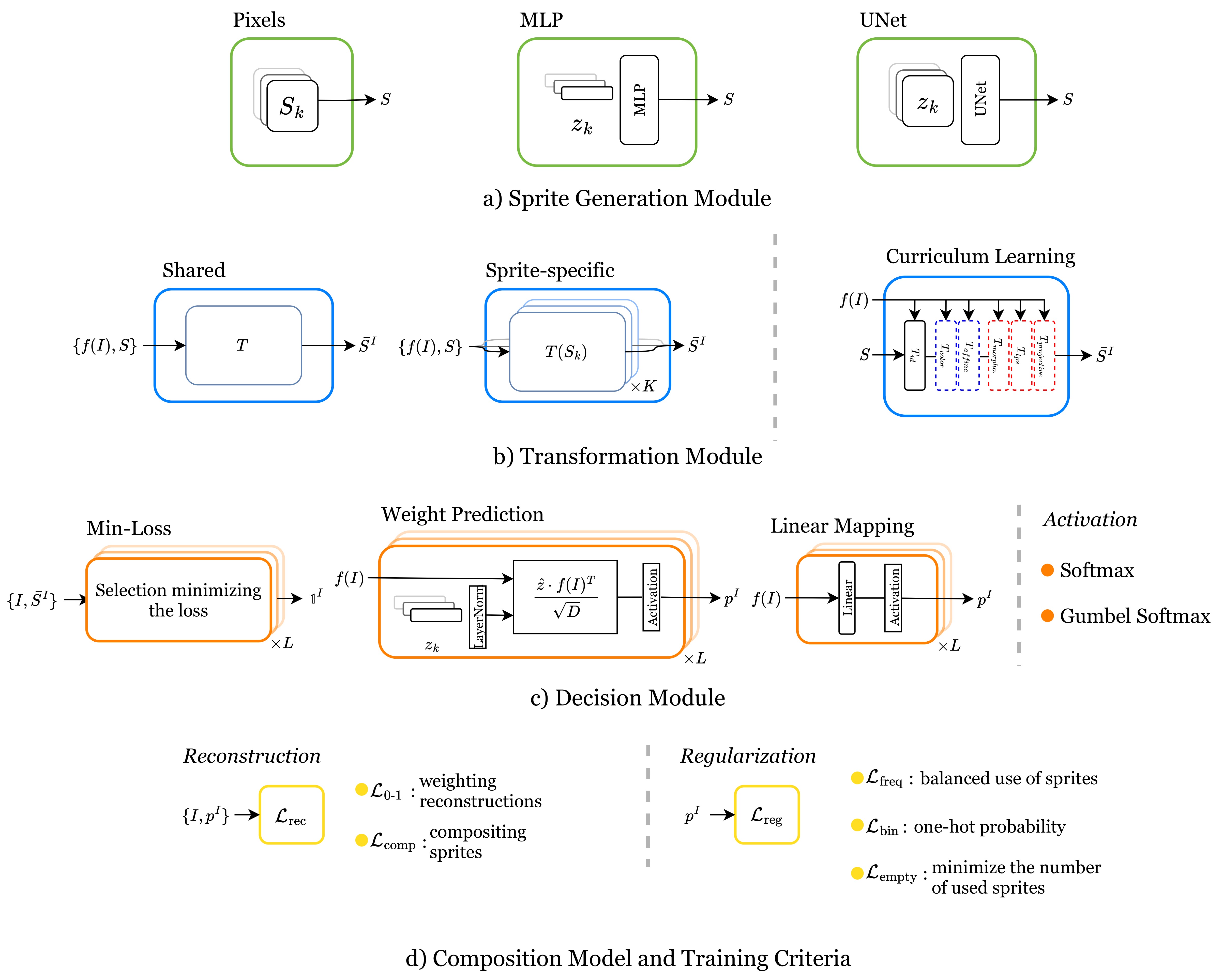}
  \caption{\textbf{Possible design choices} for the main \element{}s identified in~\cref{fig:arch}. Modules can take as input the input image $I$, features from the input image $f(I)$, the sprites $S$, the transformed sprites $\bar{S}^I$ and the predicted sprite probabilities $p^I$. (a) The \textit{Sprite Generation Module} (\greencircle{}) can learn the sprites directly as learnable parameters (\textit{Pixels}), generate them from learnable latent variables with a multi-layer perceptron (\textit{MLP}) or a \textit{UNet} architecture. (b) The \textit{Transformation Module} (\bluecircle{}) parameters can be learned with a \textit{shared} or \textit{sprite-specific} network, and with different \textit{curriculum learning} strategies. (c) The \textit{Decision Module} (\orangecircle{}) can select sprites leading to the minimum reconstruction error (\textbf{\textit{Min-Loss}}), or predict them using the sprites' latent representations (\textbf{\textit{Weight Prediction}}), or directly a linear projection (\textbf{\textit{Linear Mapping}}), with alternative activations. (d) The \textit{Composition Model and Training Criteria} (\yellowcircle{}), where the main loss can either be the sum of the reconstruction errors obtained with all the possible sprites selection weighted by their probability ($\mathcal{L}_{0-1}$) or the reconstruction error with composite sprites ($\mathcal{L}_\text{comp}$). It can also include regularizations ($\mathcal{L}_\text{\{freq, bin, empty\}}$).}
  \label{fig:modules}
\end{figure*}

In more detail, we separate sprite-based architectures into their key \element{}s, visualized in~\cref{fig:arch}: Sprite Generation Module, Transformation Module, Decision Module, and Training Criteria. For each \element{}s, we identify different design choices proposed in the literature, as well as simpler baselines, detailed in~\cref{fig:modules}. We explain how the training criteria correspond to different image composition models and are related to the exponential cost of some sprite-based image decomposition approaches. We show that one can effectively study the impact of most design choices for clustering, where the benchmarks are more realistic and diverse than for image decomposition, where they are mainly synthetic. 

Our key insight is that the main challenge of sprite-based approaches lies in jointly learning and selecting the sprites. K-means-style optimization for sprite selection leads to the discovery of more visually coherent, precise, and semantically accurate objects, without the need for complex regularization, as regularization is implicitly enforced through cluster reassignment policies.
However, this type of optimization scales exponentially with the number of objects per image. We show that different regularization techniques can improve approaches that directly predict sprite selection.
\edit{While we perform most of our analyses on the more diverse and less computationally demanding clustering benchmarks, this actually enable us to design an approach which we demonstrate can generalize to multi-layer decomposition.}

This paper is organized as follows: First, in~\cref{sec:related}, we review the literature on clustering and image decomposition. Second, in~\cref{sec:method}, we present a unified formalization for sprite-based image decomposition models. Third, in~\cref{sec:analysis_clustering}, we perform a comparative analysis of the different design choices on clustering and propose our new approach. Finally, in~\cref{sec:analysis_decomp}, we extend and evaluate our approach for multi-layer image decomposition.

\textbf{Contributions.} Our contributions are as follows:
\begin{itemize}[noitemsep]
    \item We perform an exhaustive analysis of sprite-based methods and identify their key \element{}s. 
    \item We systematically study their impact on clustering benchmarks.
    \item We propose a novel sprite-based approach that predicts sprite selection and scales linearly with the number of objects per image.
\end{itemize}

%% file: Sections/2-RelatedWork.tex
\section{Related Work}
\label{sec:related}
\subsection{Image Clustering}
We focus on image clustering approaches that are most related to our work and classify them into pixel-based and deep-feature-based clustering. For a broader literature review, we refer the reader to dedicated surveys~\citep{zhou2024comprehensive,ren2024deep,wei2024odc}. 
\subsubsection{Pixel-based Clustering}
Clustering in pixel space is highly challenging since the image content can be associated with different backgrounds and  can undergo spatial and color transformations that completely change its pixel representation. Traditional clustering methods, such as K-means~\citep{macqueen1967some}, therefore lead to limited results when applied directly on full images. EM-based transformation-invariant clustering algorithms have been proposed to gain invariance to user-defined families of transformation~\citep{frey99em,frey01fltic,frey03tic}. They operate directly in image space, compare pixel values, and provide prototypical representations of clusters. The idea of transformation invariance was also adopted in congealing-based image alignment models that learn transformations using a data-driven approach~\citep{cox2008lsc,cox2009lsc,huang07uja,miller00shared,annunziata2019jointly,lm06data}, some with a focus on clustering~\citep{mattar2012unsupervised,liu2009simal}. Deep Transformation-invariant (DTI) Clustering builds on this idea but optimizes prototypes and transformations in a deep learning framework~\cite{monnier2020dticlustering}. The sprite-based image models we study are very related to DTI-Clustering, which can be seen as a single-layer image model, where sprites correspond to prototypes. 
\subsubsection{Deep Features and Clustering}
Many recent deep architectures adopt clustering as an objective for representation learning, e.g., ~\citet{caron2018deep,caron2020unsupervised,li2021prototypical,liang2023clusterfomer}, without specifically targeting clustering performance. More relevant to us are those that specifically target clustering, focusing on various technical tools, such as CNNs~\citep{yang2016joint,chang2017deep}, autoencoders~\citep{xie2016unsupervised,mrabah2019adversarial,dizaji2017deep,kosiorek2019stacked,shaham2018spectralnet}, mutual information~\citep{hu2017learning,ji2019invariant}, generative models~\citep{jiang2016variational,mukherjee2018cluster}, or instance discrimination~\citep{niu2022spice, van2020scan}. 
The crucial and common aspect of these deep clustering approaches is relying on abstract image features. However, relying on deep representations of images and clusters in feature space makes it very hard to interpret the results, performance, and failures, especially in a visually intuitive way.
\subsection{Image Decomposition}
Image decomposition is a broad concept and could encompass broad areas of research from image co-segmentation to layered video representations. In this section, we focus on the approaches that are the most relevant to our work and are often referred to as unsupervised multi-object segmentation approaches or deep object-centric image decomposition methods. We only review single-image methods, and do not dive into the many works that leverage motion, video, or 3D. 
We follow the taxonomy of~\citet{karazija2021clevrtex}, differentiating pixel-based, glimpse-based, and sprite-based approaches. Another view of these approaches is presented in~\citet{greff2020binding}, which differentiates approaches depending on the type of slot they rely on, namely instance slots, sequential slots, spatial slots, and category slots. {For a broader review of unsupervised object discovery approaches, we refer the reader to~\citet{villa2024unsupervised}.}

\paragraph{Pixel-based methods.} Pixel-based methods assign each pixel to an image component, typically by performing probabilistic pixel clustering. Early works tackle this clustering problem by developing approaches based on denoising autoencoders (DAE)~\citep{greff2015binding,vincent2008extracting}, Iterative Amortized Grouping~\citep{greff2016tagger}, and Neural Expectation Maximization~\citep{greff2017neural}.  
However, these pioneer methods were limited to simple images with a small number of objects.

More recent models following this pixel-based paradigm include MONet~\citep{burgess2019monet}, IODINE~\citep{greff2019multiobject}, eMORL~\citep{emami2021efficient}, and GENESIS~\citep{engelcke2020genesis,engelcke2021genesis}, and demonstrate results on the more challenging synthetic CLEVR dataset of rendered 3D spheres, cubes, and cylinders. They typically output a segmentation mask for each image component as well as a latent code that enables generating an appearance image for each component. The ECON~\citep{kugelgen2020towards} method is built on MONet but is more related to our work because it explicitly models object layers and is designed to completely model occluded objects. However, it has only been demonstrated on very simple synthetic data.


Moving away from probabilistic scene representation and pixel clustering of these so-called scene-mixture models,~\citet{locatello2020objectcentric} proposed a discriminative approach to scene component identification. Their slot attention mechanism localizes scene components through an iterative clustering-like attention mechanism and leads to a latent representation for each slot, which encodes both its mask and appearance. This approach has been successfully applied to perform object discovery on much more challenging images. DINOSAUR~\citep{seitzer2023bridging} first demonstrated results on real-world datasets by applying slot attention to DINO features~\citep{caron2021emerging}, instead of pixels. It has also been combined with more complex slot decoders, including auto-regressive~\citep{singh2022illiterate,kakogeorgiou2024spot} 
and diffusion~\citep{jiang2023object,wu2023slotdiffusion,singh2025glass} ones. \\

\paragraph{Glimpse-based methods.} Glimpse-based methods first extract regions of the image containing objects and then predict object models for each region. This idea was introduced by the Attend, Infer, Repeat approach (AIR)~\citep{eslami2016attend}, which was the main inspiration for a series of works, such as SQAIR~\citep{kosiorek2018sequential}, SPAIR~\citep{crawford2019spatially}, or SuPAIR~\citep{stelzner2019faster}. Similar to early pixel-based approaches, these works were developed for very simple synthetic datasets. More recent works, such as SPACE~\citep{lin2020space}, GNM~\citep{jiang2020generative}, and AST~\citep{sauvalle2023unsupervised}, extend glimpse-based approaches to CLEVR-like datasets. However, they seem to be out-shone by slot-attention-based approaches, which brought comparable efficiency to pixel-based approaches. 


\paragraph{Sprite-based methods.} Sprite-based methods learn a set of object prototypes, referred to as sprites, and how to combine sprites to reconstruct images. These sprites make their image model much more tangible than pixel and glimpse-based approaches, enabling them to discover object categories, not just instance segmentation. StampNet~\citep{visser2019stampnet} can be considered as the first deep sprite-based approach. It learns a latent space to categorize and localize objects, but was only demonstrated on very simple synthetic datasets. Capsule approaches~\citep{kosiorek2019stacked,xiang2021dpr} are similar in spirit and have been categorized as sprite-based methods in~\citet{villa2024unsupervised}. However, they learn abstract feature-based representations of object parts, they are typically evaluated on clustering benchmarks, and to the best of our knowledge, they have not been demonstrated on standard multi-object datasets.
Appealing results on video-game and text images have been demonstrated by MarioNette~\citep{smirnov2021marionette}, which sees sprite discovery as a self-supervised learning problem, and learns to predict sprite occurrence and position. DTI-Sprites~\citep{monnier2021unsupervised} models sprite shape and color transformation, enabling it to tackle more complex datasets, including CLEVR data. However, it requires testing many sprite configurations instead of predicting sprite occurrence, and thus does not scale to a large number of objects. Focusing on text line analysis, the Learnable Typewriter~\citep{siglidis2024ltr} combines ideas from MarioNette and DTI-Sprites for applications in digital humanities. Similar ideas have been applied to model 3D point clouds~\citep{loiseau2023learnable}.  Our analysis encompasses all these sprite-based approaches and clarifies their differences.  

%% file: Sections/3-Method.tex
\section{Sprite-based Approaches}
\label{sec:method}
\begin{table*}[t!]
    \caption{\textbf{Comparison of sprite-based models.} Existing sprite-based methods make very different choices for several of the four components that we identified, making direct comparison between their performance difficult. Our exhaustive analysis leads to an informed choice of all the components for clustering (Ours-C) and image decomposition (Ours-D).}
    \label{tab:module-setup}
    \centering
    \tabcolsep=1pt
    \resizebox{\linewidth}{!}{
    \begin{tabular}{@{}lcccccccc@{}}
    \toprule
        Method & \greencircle{} Sprite Generation  & \bluecircle{} Curriculum & \bluecircle{} Sharing & \orangecircle{} Decision & \orangecircle{} Activation & \yellowcircle{} $\mathcal{L}_\text{rec}$ &  \yellowcircle{} $\mathcal{L}_\text{reg}$ & \faNetworkWired \\
        \midrule
        StampNet~\cite{visser2019stampnet}& pixels & all & sprite-specific & linear & hard GS & $\mathcal{L}_\text{comp}$& - & decomposition  \\
        DTI-Clustering~\cite{monnier2020dticlustering} & pixels & one-by-one & sprite-specific & Min-Loss & none & $\mathcal{L}_{\text{comp} =0-1}$ & reassignment  & clustering \\
        DTI-Sprites~\cite{monnier2021unsupervised} & pixels& one-by-one  & sprite-specific & Min-Loss & softmax & $\mathcal{L}_{\text{comp} =0-1}$ & reassignment, $\mathcal{L_\text{empty}}$  & decomposition  \\
        MarioNette~\cite{smirnov2021marionette} & MLP & all& shared  & weight prediction & none & $\mathcal{L}_\text{comp}$ & $\mathcal{L}_\text{bin}$ & decomposition  \\
        Learnable Earth Parser~\cite{loiseau2023learnable} & 3D point clouds & one-by-one & shared & linear & softmax & $\mathcal{L}_{0-1}$ &  $\mathcal{L}_\text{freq}$ & decomposition \\
        Learnable Typewriter~\cite{siglidis2024ltr} & MLP & all & shared & weight prediction & softmax & $\mathcal{L}_\text{comp}$ & - & decomposition  \\
        \textbf{Ours-C} & MLP & one-by-one & sprite-specific & linear & soft GS & $\mathcal{L}_\text{comp}$ & $\mathcal{L}_\text{\{freq,bin\}}$ & clustering \\
        \textbf{Ours-D} & MLP & one-by-one & sprite-specific & linear & soft GS & $\mathcal{L}_\text{comp}$ & $\mathcal{L}_\text{empty}$ & decomposition \\
    \bottomrule
    \end{tabular}}
\end{table*}
In this section, we first present a unified view of sprite-based decomposition methods for clustering and layered image decomposition, summarized in \figref{fig:arch}. Then, for each key \element{} that we identify, we detail different design choices that have been introduced in the literature and that we consider in our study, which are summarized in \figref{fig:modules}, and are related to the literature in~\cref{tab:module-setup}.

\subsection{Unified View and Formalization}
Our key insight is that sprite-based approaches rely on four main \element{}s that we present first. We then discuss how these \module{}s can be used for multi-object image decomposition and clustering. The choices made by different sprite-based approaches in the literature are summarized in~\cref{tab:module-setup}.

\subsubsection{Key Components}

Sprite-based approaches take as input an image $I\in \mathbb{R}^{W\times H \times C}$, with $C=1$ for a grayscale image and $C=3$ for an RGB image, and predict a set of layers associated with this image. As visualized in \figref{fig:arch}, we identified four key \element{}s in sprite-based methods:
\begin{itemize}
    \item A {\bf sprite generation \module{}}, $G$ (\secref{subsec:sprite}), which learns $K$ sprites $S_{1},\cdots,S_{K}$, with for all $ k\in \{1,\ \cdots, K\}$,  $S_{k}\in \mathbb{R}^{R\times R \times C'}$, where $R$ is the size of the sprites and $C'$ is the number of channels per sprite. Sprites can be interpreted as prototypical images and can include segmentation, encoded as a transparency mask. Depending on the approach, $C'$ can be 1 (a grayscale image), 2 (a grayscale image and transparency), 3 (an RGB image), or 4 (an RGB image with a transparency channel).
    \item A {\bf transformation \module{}}, $T$ (\secref{subsec:tr}), which takes as input a target image $I$ and sprites $S_{1},\cdots,S_{K}$, and outputs a set of transformed sprites $\bar{S}^I$. Transformation typically includes color and spatial transformations. The transformed sprites are images of the same size as the input image $I$, with an optional transparency channel. Note that this \module{} can predict several transformations for each sprite, enabling the modeling of images with multiple elements, as we clarify in~\cref{sec:decompo}.
    \item A {\bf decision \module{}}, $P$ (\secref{subsec:decision}), which predicts probabilities $p^I$ for each of the transformed sprites to be used in the reconstruction of the input image. %
    \item A {\bf reconstruction loss}, $\mathcal{L}$ (\secref{subsec:loss}), which evaluates how well the transformed sprites associated with the predicted probabilities explain the input image, and with which the model is optimized.
\end{itemize}
These \element{}s and the losses correspond to an image formation model, $\mathcal{C}(\bar{S}, p)$. 

\subsubsection{Layered Image Decomposition}
\label{sec:decompo}

For layered image decomposition, one typically assumes a maximum number of layers $L$. Each sprite $S_k$ for $k\in \{1,\ \cdots, K\}$ is then transformed into $L$ sprites $\bar{S}^I_{k,l} \in\mathbb{R}^{W\times H \times C'}$ for $l\in \{1,\ \cdots, L\}$, with $C'=C+1$, leading to a set of $K\times L$ transformed sprites $\bar{S}^I=(\bar{S}^I_{1,1},\cdots,\bar{S}^I_{K,L})$. Sprites are selected according to $p^I\in[0,1]^{K\times L}$. Note that one of the sprites can be used as an empty sprite, \ie, frozen and completely transparent, to allow modeling a variable number of objects. Background can be modeled using one or several specific opaque sprites, possibly with particular constraints (e.g., having a uniform color) and be associated with their own specific transformations. To simplify notation, we do not differentiate background sprites from the other sprites. In our experiments on layered image decomposition, we always model the background with a single sprite.  
The image formation model, $\mathcal{C}$, composites the transformed background sprite with the sprites from the following layers. 
\edit{To better handle occlusion, we follow DTI-Sprites~\citep{monnier2021unsupervised} and predict a matrix defining the order of the layers.}

\subsubsection{Clustering}
In the case of clustering, the simplest scenario~\citep{monnier2020dticlustering} is to consider a single-layer image model using only completely opaque sprites.
In that case, the set of transformed sprites is $\bar{S}^I=(\bar{S}^I_{1,1},\cdots,\bar{S}^I_{K,1})$, with for all $k\in \{1,\ \cdots, K\}$, $\bar{S}^I_{k,1} \in\mathbb{R}^{W\times H \times C}$ the transformed version of sprite $S_k$. Note that if there are no transformations, and the $L_2$ loss between the input and the transformed sprite that best approximates it is optimized, this model boils down to standard K-means~\citep{macqueen1967some,bottou1994convergence}.

Another approach that typically leads to better results for more complex images~\citep{monnier2021unsupervised} is to explicitly model the background using a background sprite and the different clusters with sprites including a transparency channel, and thus consider a 2-layer model. 
The image formation model, $\mathcal{C}$, composites the transformed background sprite with the other transformed sprites depending on the output $p^I\in[0,1]^K$ of the selection \module{}. 

Both of these approaches can be seen as specific cases of layered image decomposition and leverage the same modules, enabling us to start our analysis by focusing on the simpler clustering scenario.

\subsection[Sprite Generation]{%
    \greencircle{} Sprite Generation \Module{}%
}
\label{subsec:sprite}
The sprites $S_1,\cdots,S_K$ are the visual representation of the recurrent patterns identified by the model in the target image collection. They are thus common to all input images $I$, they are themselves modeled as images -- color or grayscale, and associated or not with a transparency mask -- and they can be learned with different strategies.
\subsubsection{Learning Pixel Values}
\label{subsubsec:pixel}
Directly learning the sprite, \ie setting each sprite's pixel values as learnable parameters, is the simplest choice and has been used in~\cite{monnier2020dticlustering,monnier2021unsupervised}. 
\subsubsection{Decoding Learned Latent Variables with a Generator Network (MLP or U-Net)}
\label{subsubsec:latent}
Motivated by the possibility of using latent variables to link sprite generation and clustering,~\citet{smirnov2021marionette} proposes to learn $K$ latent vectors $z_{1},\cdots,z_{K}$ and a generation network ${G}$ that takes as input those latent vectors, and outputs the corresponding sprite $S_k={G}(z_k)$. Note that while generated by a network, the sprites still do not depend on the input image $I$, and that once the network is trained, they could be computed once and for all, without using the generator network. Following~\citet{siglidis2024ltr}, we explore the use of a Multi-Layer Perceptron (MLP) or a U-Net architecture~\citep{ronneberger2015u} (U-Net) as the generation network. 

\subsection[Transformation]{%
    \bluecircle{} Transformation \Module{}%
}
\label{subsec:tr}
Sprite-based approaches account for variations in the appearance of objects in terms of shape or color by explicitly modeling them. Given an input image $I$, they predict one (for clustering) or several (for image decomposition) transformations for each sprite. The family of transformations that are available and the way in which they are learned are important hyperparameters, and the optimal choice depends on the target dataset. Transformations typically include (i) spatial transformations, modeled with Spatial Transformer Networks~\citep{jaderberg2015spatial}, and (ii) affine color transformation, where parameters are predicted and applied on the sprite values. They may include more specific transformations, such as morphological transformations to model stroke width for the MNIST dataset~\citep{lecun2010mnist}. There are several key design choices in this transformation learning that we explore.

\blockcomment{\begin{figure}[t]
\centering
\includegraphics[width=0.9\columnwidth]{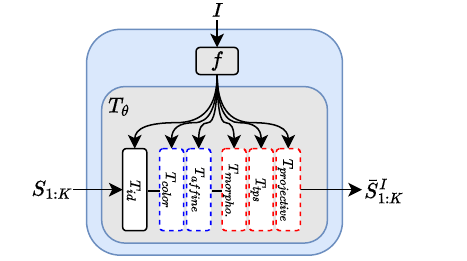}
\caption{\bluecircle{}\textbf{Details of the Transformation \Module{}.} Orders and groupings of transformations, including coarse transformations (in blue): identity, color, and affine, as well as fine transformations (in red): morphological, TPS, and projective. We denote each \color{blue}{`g1'}\color{black} and \color{red}{`g2'}\color{black}, respectively.}
\label{fig:tr_curr}
\end{figure}}

\subsubsection{Curriculum Learning}
Because transformations could model dramatic changes, curriculum learning is the key to progressively learning meaningful transformations. We explore various curriculum scheduling strategies. To study them, we first decided on a fixed order of transformation by increasing complexity, as visualized in~\figref{fig:modules}b: no transformation, affine color transformation, affine spatial transformation, morphological transformation, Thin Plate Spline (TPS) transformation, and projective transformation. 
With all transformations initialized as the identity function, we then tested different strategies:
\begin{itemize}[noitemsep] 
    \item \textbf{\textit{all}}:  optimizing all transformations together from the start,
    \item \textbf{\textit{id+rest}}: learning first without any transformation, then optimizing all transformations together,
    \item \textbf{\textit{id+g1+g2}}: grouping transformations into three groups -- (id) no transformation, (g1) affine color and spatial transformations, (g2) other transformations -- and adding each group of transformations into the optimization one-by-one, and
    \item \textbf{\textit{one-by-one}}: adding each of them into the optimization one-by-one.
\end{itemize}
Note that for each dataset, we only use transformations relevant to the dataset (see Appendix~\cref{tab:tr-selection}).
\subsubsection{Sprite-Specific vs. Shared Transformations}
Another question we explore is the possibility and consequences of sharing the transformations among sprites. Intuitively, one could expect the sprites to be better aligned if the same transformations are applied to all sprites, while an architecture that applies specific transformations to all sprites might be more powerful. Sharing transformations might also be beneficial when modeling a large number of sprites.
\subsection[Decision]{\orangecircle{} Decision \Module{}}
\label{subsec:decision}
A crucial problem of sprite-based approaches is deciding which transformed sprites to use to reconstruct a specific image. We consider two types of solutions.
\subsubsection{Minimum Loss}
A simple approach is to choose the sprites that minimize the loss~\citep{bottou1994convergence,monnier2020dticlustering}. However, this means that (i) during training, only sprites that are selected receive gradients, and thus some might never be used, which requires specific re-assignment strategies, and (ii) when modeling images with multiple objects, the number of possible sprite combinations is exponential in the number of objects, which complicates optimization.  Note that this approach can be seen as a deterministic layer predicting one-hot probability vectors $p^I$, and we refer to it as \textbf{\textit{Min-Loss}}.
\subsubsection{Probability Prediction}
Another approach is to use a neural network to predict which transformed sprites should be used for a specific input image $I$ by predicting probability distributions among transformed sprites. While this is much more in line with common deep learning paradigms, we show experimentally that jointly learning the sprites, their transformations, and the selection of the best sprites is a challenging optimization problem, which requires using many regularization functions that make the method more specific and less robust. 

The more standard architecture to predict such a probability distribution is a network that takes as input the target image $I$ and finishes with a linear layer and a softmax, which we refer to as \textbf{\textit{linear mapping}}. However, MarioNette~\citep{smirnov2021marionette} proposes having a network instead predict classification weights from latent variables, shared with the sprite generation \module{}, which are then compared with the input image features, before applying a softmax. We refer to this approach as \textbf{\textit{weight prediction}}.

Finally, because what is ultimately needed is a binary selection of the sprites, we experimented with replacing the softmax by Gumbel softmax~\citep{jang2017categorical,maddison2017concrete}, similar to StampNet~\citep{visser2019stampnet}. However, while StampNet uses Gumbel softmax with binary selection, we use Gumbel softmax with soft selection, which consistently led to better performances.

\subsection[Composition Model and Training Criteria]{\yellowcircle{} Composition Model and Training Criteria}
\label{subsec:loss}
We decompose the training loss as a reconstruction loss, $\mathcal{L}_{\text{rec}}$, and a  regularization loss, $\mathcal{L_{\text{reg}}}$:
\begin{equation}
    \mathcal{L} = \mathcal{L}_{\text{rec}} +\mathcal{L}_{\text{reg}} ~.
    \label{eq:final-loss}
\end{equation}
We study two reconstruction losses, which actually correspond to two different composition models $\mathcal{C}(\bar{S}, p)$, as well as different regularization losses.

\subsubsection{Composition Model and Reconstruction Loss}
The composition model, $\mathcal{C}(\bar{S}^I, p^I)$, composites transformed sprites into an image. The first way to see this model is to consider that it can only select sprites in a binary way, and thus, the loss should be a weighted sum of reconstruction errors of each sprite selection weighted by their probability ($\mathcal{L}_{\text{0-1}}$). The second way to build composite sprites is by weighting transformed sprites according to predicted probabilities $p$, then reconstructing images with composite sprites ($\mathcal{L}_{\text{comp}}$).%


More formally, let us define $\mathcal{C}^L$ the standard alpha-blending composition of $L$ images $(A_1,\alpha_1),\cdots,(A_L,\alpha_L)$, where for $l=1,\cdots , L$ the $A_l$ are RGB images and $\alpha_l$ their associates transparency channels:
\begin{equation}
    \mathcal{C}^L\left(\left(A_1,\alpha_1\right),\cdots,(A_L,\alpha_L) \right)= \sum_{l=1}^L  \left( \alpha_l \prod_{k=l+1}^L (1-\alpha_k)\right)A_l,
\end{equation}
where the product is 1 if empty and multiplications are to be understood pixelwise.
Let us consider probabilities $p^I \in \mathbb{R}^{K\times L} $ and transformed sprites $\bar{S}^I_{k,l} \in\mathbb{R}^{W\times H \times C}$ for all $k\in \{1,\ \cdots, K\}$ and $l\in \{1,\ \cdots, L\}$. Then $\mathcal{L}_{\text{0-1}}$ is defined by:
\begin{equation}
\mathcal{L}_{\text{0-1}}(\bar{S}^I,p^I)=\sum_{(k_1,\cdots k_L)\in \{0,\cdots,K\}^L} 
\left( \prod_{l=1}^L p^I_{k_l,l} \right) ||I-\mathcal{C}^L(\bar{S}^I_{k_1,1},\cdots,\bar{S}^I_{k_L,L})||_2^2 ~,
\label{eq:loss-diff}
\end{equation}
and $\mathcal{L}_{\text{comp}}$ is defined by:
\begin{equation}
\mathcal{L}_{\text{comp}}(\bar{S}^I,p^I)=||I-\mathcal{C}^L(\sum_{k=1}^K p^I_{k,1} \bar{S}^I_{k,1},\cdots,\sum_{k=1}^K p^I_{k,L} \bar{S}^I_{k,L} )||_2^2 ~.
\label{eq:loss-sprite}
\end{equation}

As can be seen from the equations, $\mathcal{L}_{\text{0-1}}$ requires to compute $K^L$ composite images, which is computationally prohibitive for large numbers of sprites and layers, while $\mathcal{L}_{\text{comp}}$ only requires computing $L$ composed sprites and a single composite images. However, $\mathcal{L}_{\text{comp}}$ corresponds to an image composition model where different transformed sprites can be merged, which might lead to undesired optima where objects are represented by overlapping multiple sprites. 
Note that in the case where $p^I$ is binary, we have $\mathcal{L}_{\text{0-1}}=\mathcal{L}_{\text{comp}}$.

%
%
\subsubsection{Regularizations}

We consider three regularization losses.

First, $\mathcal{L}_{\text{freq}}$ attempts to prevent some sprites from never being used, by penalizing using a sprite with a frequency lower than a scalar value $\epsilon \in [0,1]$:
\begin{equation}
    \mathcal{L}_{\text{freq}} = \sum_{k=1}^{K} \max \left( 0, \epsilon - \frac{1}{|D|} \sum_{I} \sum_{l=1}^{L} p_{k,l}^I \right)~,
    \label{eq:loss_reg_freq}
\end{equation}
where in practice the loss is computed over a batch of images $I$. 
Note that DTI-Sprites~\citep{monnier2021unsupervised} has a re-assignment strategy for unused sprites that plays a similar role and has a similar minimum frequency hyperparameter $\epsilon$. 

%

Second, $\mathcal{L}_{\text{bin}}$ encourages one-hot probability vectors $p^I$, and thus attempts to avoid several transformed sprites being used together to reconstruct an object. Thus, it is particularly meaningful to regularize $\mathcal{L}_{\text{comp}}$. Following~\citet{smirnov2021marionette}
, we define $\mathcal{L}_{\text{bin}}$ by:
\begin{equation}
    \mathcal{L}_{\text{bin}} = \frac{1}{K}\sum_{k=1}^K\text{Beta}(2,2)\left(p_{k}^I\right) ~,
    \label{eq:loss_reg_bin}
\end{equation}
where the probability density function of the Beta distribution is given by:
\[
f(x; \alpha, \beta) = 
\begin{cases} 
\frac{x^{\alpha-1} (1-x)^{\beta-1}}{B(\alpha, \beta)} & \text{for } 0 < x < 1 \\
0 & \text{otherwise}
\end{cases}
\]
where \( \alpha > 0 \) and \( \beta > 0 \) are the shape parameters, and \( B(\alpha, \beta) \) is the Beta function, defined as:

\[
B(\alpha, \beta) = \int_0^1 t^{\alpha-1} (1-t)^{\beta-1} \, dt = \frac{\Gamma(\alpha) \Gamma(\beta)}{\Gamma(\alpha + \beta)},
\]

where, \( \Gamma(\cdot) \) is the Gamma function, which generalizes the factorial function $(n-1)!$.

%
{Third, following~\citet{monnier2021unsupervised}, $\mathcal{L_\text{empty}}$ encourages the model to use as few sprites as possible to reconstruct an image, and attempts to avoid failure cases like sprites used with a high transparency to better reconstruct details of the images. It penalizes the use of non-empty sprites, and writing $e$ the index of the empty sprite can be defined as:
\begin{equation}
\mathcal{L}_\text{empty} = 
\sum_{l=1}^{L}
(1-p^I_{e,l}) ~.
\label{eq:empty}
\end{equation}

$\mathcal{L}_{\text{reg}}$ is defined as a weighted sum of these three regularization losses:
\begin{equation}
\mathcal{L}_{\text{reg}} = \lambda_{\text{freq}} \mathcal{L}_{\text{freq}} + \lambda_{\text{bin}} \mathcal{L}_{\text{bin}}+ \lambda_{\text{empty}} \mathcal{L}_{\text{empty}} ~,
\end{equation}
with $\lambda_{\text{freq}}$, $\lambda_{\text{bin}}$  and $\lambda_{\text{empty}}$  scalar hyperparameters.

%% file: Sections/4-Analysis.tex
\section{Analysis on Clustering}
\label{sec:analysis_clustering}
In this section, we analyze single-layer sprite-based approaches on clustering, for which experiments are faster, and datasets are more diverse than for unsupervised image decomposition, and we  leverage this analysis to define a new approach for sprite-based clustering. \cref{subsec:exp_setup} introduces the details of our experimental setup. \cref{subsec:res_sprite,subsec:res_tr,subsec:res_dec} present comparative analysis of approaches through experiments on \textit{Sprite Generation}, \textit{Transformation}, \textit{Decision}, and \textit{Training Criteria} \edit{(see \cref{fig:modules} for an overview and terminology)}. Finally, \cref{sec:clustering_comp} compares our clustering approach with the state-of-the-art. 
\subsection{Experimental Setup}
\label{subsec:exp_setup}
\subsubsection{Datasets}
We conducted experiments on 8 datasets with different characteristics: MNIST~\citep{lecun2010mnist}, ColoredMNIST~\citep{arjovsky2019invariant}, FashionMNIST~\citep{xiao2017fashion}, AffNIST~\citep{affnist}, USPS~\citep{hull1994usps}, FRGC~\citep{phillips2005frgc}, SVHN~\citep{netzer2011reading}, and GTSRB-8~\citep{stallkamp2012man} (detailed in the~\cref{sec:supmat_data}).
Digit datasets~\citep{lecun2010mnist,arjovsky2019invariant,hull1994usps,netzer2011reading} differ in complexity, ranging from grayscale digits to real-world RGB street number images. 
The other datasets tackle fashion items~\citep{xiao2017fashion}, faces~\citep{phillips2005frgc}, and traffic signs~\citep{stallkamp2012man}, offering a diversity of challenges.

\subsubsection{Training Details and Evaluation}
We report the mean accuracy over all samples for clustering using Hungarian matching~\citep{kuhn1955hungarian} for cluster-to-class assignments. 
To evaluate the impact of our regularization losses, we provide a sensitivity analysis in~\cref{subsubsec:hyper} over a sequential grid search for the regularization weights. This analysis shows that performance variance across the searched ranges is low, establishing that while $\mathcal{L}_\text{freq}$ is crucial to prevent cluster collapse, the model remains robust to minor hyperparameter variations. Details of the training setup and complete hyperparameter configurations are provided in~\cref{app:training}.
Unless stated otherwise, we report the mean and standard error over 10 runs for each experiment.

\subsubsection{Reference Setting} We sequentially evaluate the influence of each of the key \element{} we have identified, starting from the DTI-Clustering setting~\citep{monnier2020dticlustering}, which demonstrated competitive results for clustering, and which is closest to the K-means baseline. Note that our notion of a sprite encompasses the notion of prototype used in DTI-Clustering. Then, in each section, we define a new reference setting for each of our \element{}, depending on our experimental analysis.

\subsection[Sprite Generation]{%
    \greencircle{} Sprite Generation \Module{}%
}
\label{subsec:res_sprite}
\begin{table*}[t]
    \centering
    \caption{\greencircle{} \textbf{Analysis of the sprite generation \module{} for clustering.} We report the performances of \edit{sprite generation approaches.} 
    We report accuracy (\%) and standard error over 10 runs. \bluecircle{}: one-by-one, sprite-specific transformation, \orangecircle{}: Min-Loss, \yellowcircle{}: $\mathcal{L}_{\text{comp}=0-1} $, reassignment.}
    \resizebox{\columnwidth}{!}{
    \begin{tabular}{@{}l l c c c c c c c c c@{}}
    \toprule
        \Module{} & Init.  & MNIST & ColoredMNIST & FashionMNIST & AffNIST & USPS & FRGC & SVHN & GTSRB-8 & Average \\
        \midrule
        \multicolumn{2}{c}{\textbf{\textit{Pixel Space}}} & & & & & & & & & \\
        Pixels &sample &  \textbf{97.2±0.0} & 94.5±1.5 & \underline{58.3±0.6} &  93.3±2.0 & \underline{86.3±2.0} & \underline{40.4±0.8} & 42.8±2.4 & \textbf{51.4±1.5} & 70.5 \\
        Pixels & random & \textbf{97.2±0.0} & \textbf{95.5±1.3} & 57.2±0.7 & 89.5±2.0 & 84.0±0.5 & \textbf{40.8±0.7} & 42.2±2.0 & \underline{51.2±0.6} & 69.7 \\
        \multicolumn{2}{c}{\textbf{\textit{Latent Space}}} & & & & & & & & & \\
        MLP &\multirow{2}{*}{random} &  97.1±0.0 & 94.3±1.5 & \textbf{58.9±0.7} & \textbf{95.7±1.3} & {85.5±2.3} & 40.3±0.4 & \textbf{45.8±1.2} & 51.1±1.6 & 71.1\\
         UNet& & 97.1±0.1 & \underline{94.8±1.5} & 57.9±1.3 & \underline{94.5±1.8} & \textbf{86.6±1.5} & 33.7±1.8 & \underline{45.8±2.4} & 50.3±1.0 & 70.1 \\
    \bottomrule
    \end{tabular}}
    \label{tab:sprite-ablation-clustering}
\end{table*}
\begin{wrapfigure}{r}{0.6\columnwidth}
    \centering
    \includegraphics[width=0.6\textwidth,trim={0 1mm 0 1mm},clip]{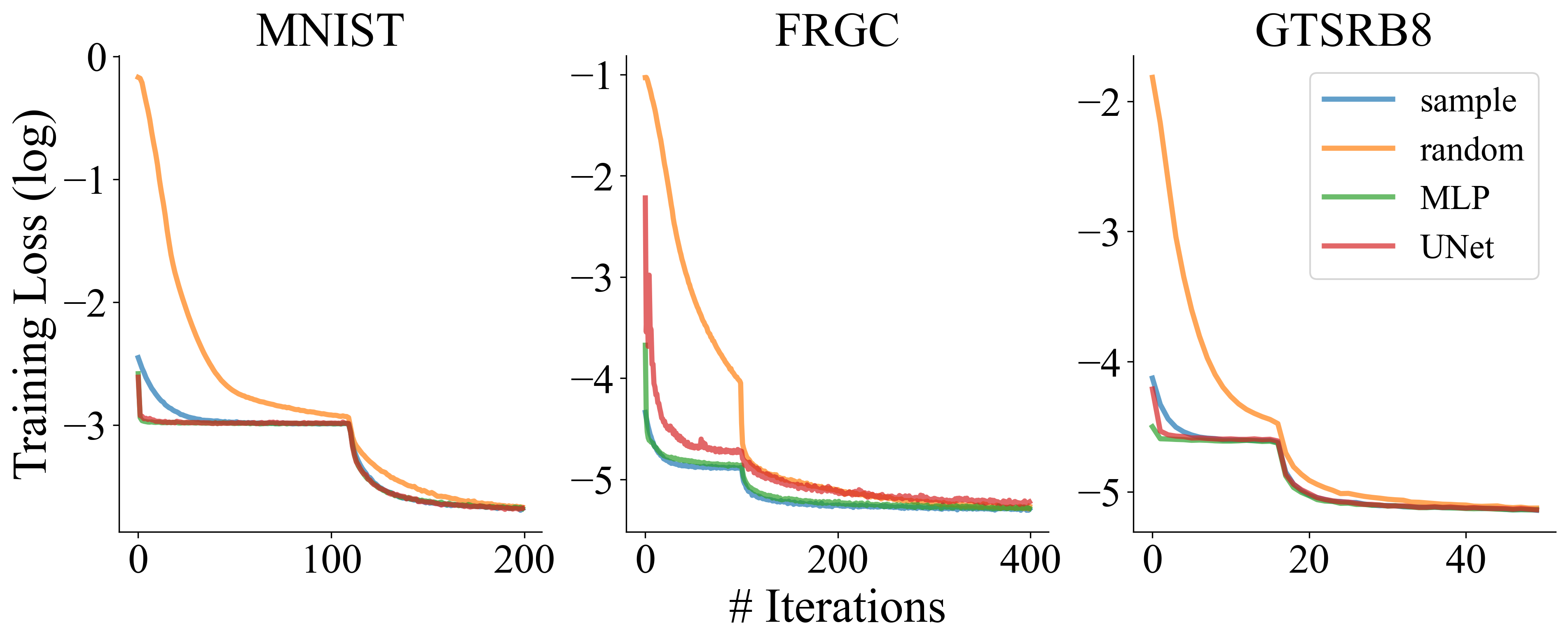}
    \caption{\greencircle{} \textbf{Training loss for different sprite generation \module{}s.} We show the average loss over 10 runs for 3 datasets. For all datasets, learning sprites through a generator network converges faster. Better seen in the digital version.} 
    \label{fig:loss-conv}
\end{wrapfigure}
As detailed in~\cref{subsec:sprite} and~\figref{fig:modules}a, we compare directly learning pixel values and learning sprites through a generator network. 
When learning pixel values, we compare initializing the sprites randomly or from a random sample, similar to the original DTI-Clustering~\citep{monnier2020dticlustering}.  
When learning sprites in latent space, we compare using a two-layer MLP and a UNet. For the MLP, we learn a latent representation of size 128 and use a hidden layer with 128 units. For UNet, we used a latent representation with the same dimension as the sample sprite and the architecture of~\cite{siglidis2024ltr}.

Our results, reported in~\cref{tab:sprite-ablation-clustering}, show that the best performing approach depends on the dataset. Learning sprites with an MLP leads to slightly better results on average\edit{, it is within or close to the error margin of the best method on all datasets, and clearly improving on pixel-based approaches on AffNIST and SVHN}. Moreover, an analysis of the training loss curves shown in \cref{fig:loss-conv} shows that learning sprites through generator networks leads to clearly faster convergence. We thus adopt \textbf{learning sprites through an MLP for the rest of our analysis.}
\subsection[Transformation]{%
    \bluecircle{} Transformation \Module{}%
}
\label{subsec:res_tr}
As explained in~\cref{subsec:tr}, we explore different constraints on the deformation \module{}, namely different curriculum and weight-learning strategies. 
\subsubsection{Curriculum Learning}
\begin{table}[ht]
    \centering
    \tabcolsep=4pt
    \caption{\bluecircle{} \textbf{Effect of curriculum learning on the transformation \module{}.} We explore different curriculum strategies. We report accuracy (\%) and standard deviation over 10 runs. \greencircle{}: MLP, \bluecircle{}: sprite-specific transformation, \orangecircle{}: Min-Loss, \yellowcircle{}: $\mathcal{L}_{\text{comp}=0-1} $, reassignment.}
    \resizebox{0.5\textwidth}{!}{
    \begin{tabular}{@{}l c c c c @{}}
        \toprule
        & \multicolumn{4}{c}{\textit{\textbf{Curriculum strategy}}} \\
        \cmidrule{2-5}
        Dataset & all & id+rest & id+{g1}+{g2} & one-by-one \\
        \midrule
        MNIST & 88.1±2.6 & 95.8±1.1 & \underline{95.8±0.9} & \textbf{97.1±0.0} \\
        ColoredMNIST & 82.8±2.7 & 83.9±2.4 & \underline{94.2±1.8} & \textbf{94.3±1.5} \\
        FashionMNIST & 56.0±1.2 & \underline{58.7±1.8} & 57.8±0.9 & \textbf{58.9±0.7} \\
        AffNIST & 83.1±4.0 & 81.7±3.0 & \textbf{95.8±1.4} & \underline{95.7±1.3} \\
        USPS & 81.3±2.1 & \textbf{86.0±1.8} & 83.2±1.2 & \underline{85.5±2.3} \\
        FRGC & 34.9±0.9 & 34.5±0.5 & \underline{39.1±0.5} & \textbf{40.3±0.4} \\
        SVHN & 32.6±2.2 & 33.3±0.4 & \textbf{45.8±1.2} & \textbf{45.8±1.2} \\
        GTSRB-8 & 49.3±1.5 & \textbf{51.3±1.8} & \underline{51.1±1.6} & \underline{51.1±1.6}\\
        Average & 63.5 & 65.7 & 70.4 & 71.1 \\
        \bottomrule
    \end{tabular}}
    \label{tab:tr-curr}
\end{table}
In~\cref{tab:tr-curr}, we report results using various curriculum strategies to learn the transformations presented in~\cref{subsec:tr}. They show that curriculum is critical for good performance. A 2-step-only curriculum, which can be interpreted as a K-means initialization followed by a full unfreeze of the network, is not sufficient, while splitting transformations into two groups already leads to good results. One-by-one curriculum performs best, and we thus continue using \textbf{one-by-one curriculum for the rest of our analysis.} 

\subsubsection{Shared Transformations}
\begin{table}[ht!]
    \centering
    \caption{\bluecircle{} \textbf{Effect of sharing transformations among sprites in the transformation \module{}.} We report accuracy (\%) and standard deviation over 10 runs. \greencircle{}: MLP, \bluecircle{}: one-by-one, \orangecircle{}: Min-Loss, \yellowcircle{}: $\mathcal{L}_{\text{comp}=0-1} $, reassignment.}
    \resizebox{0.5\textwidth}{!}{
    \begin{tabular}{@{}l c c}
    \toprule
        Dataset & {Shared transfo.} & {Sprite-specific transfo. } \\
        \midrule
        MNIST & 91.9±2.2 & \textbf{97.1±0.0} \\
        ColoredMNIST & 92.6±2.0 & \textbf{94.3±1.5}\\
        FashionMNIST & 57.0±0.4 & \textbf{58.9±0.7}\\
        AffNIST & 86.4±2.8 & \textbf{95.7±1.3} \\
        USPS & 84.4±2.3 & \textbf{85.5±2.3}\\
        FRGC & \textbf{41.1±0.6} & 40.3±0.4\\
        SVHN & 34.3±0.1 & \textbf{45.8±1.2}\\
        GTSRB-8 & 49.2±1.2 & \textbf{51.1±1.6}\\
        Average & 67.1 & 71.1 \\
    \bottomrule
    \end{tabular}}
    \label{tab:tr-shared}
\end{table}
\begin{figure}[ht!]
    \centering
    \begin{subfigure}{0.45\columnwidth}
        \includegraphics[width=\columnwidth]{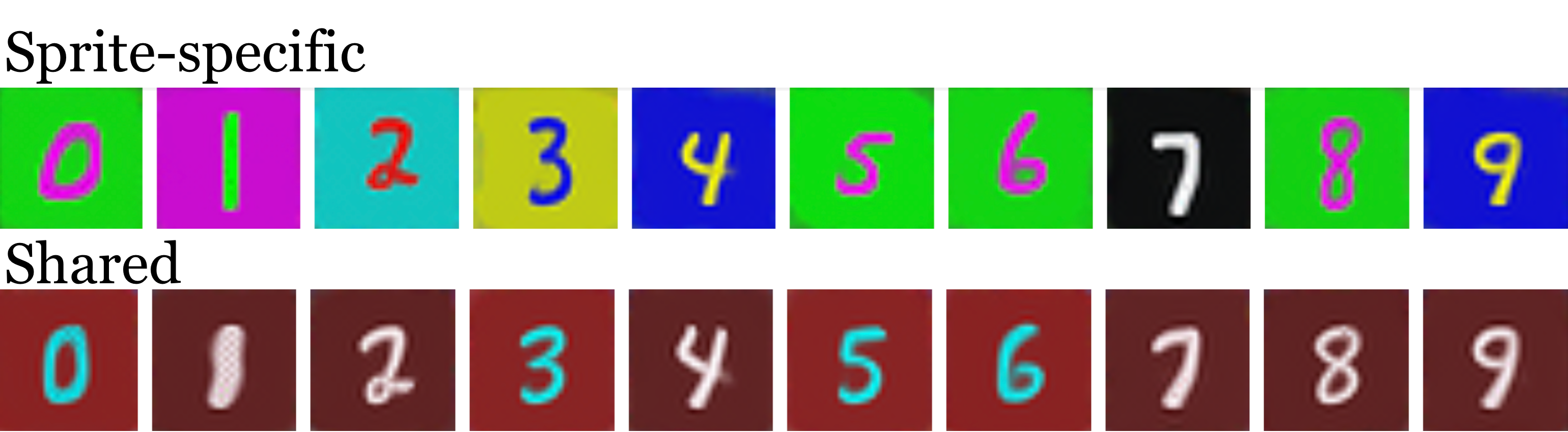}
        \caption{ColoredMNIST}
        \label{subfig:shared-tr-qual-color}
    \end{subfigure}
    \hfill
    \begin{subfigure}{0.45\columnwidth}
        \includegraphics[width=\columnwidth]{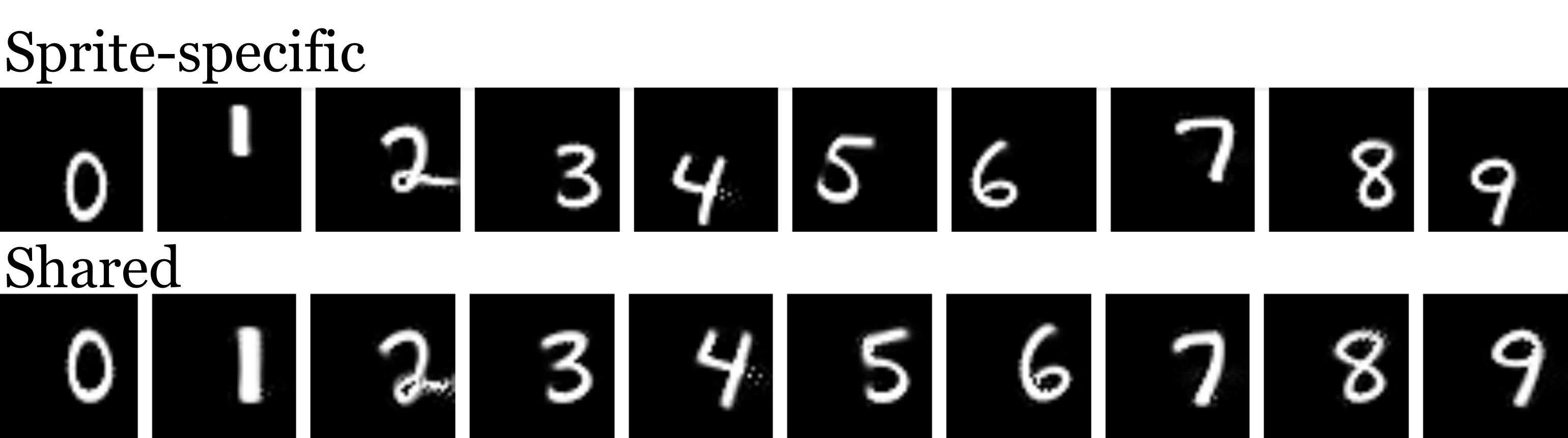}
        \caption{AffNIST}
        \label{subfig:shared-tr-qual-aff}
    \end{subfigure}
    \caption{\bluecircle{} \textbf{Qualitative effect of sharing transformations among sprites in the transformation \module{}.} We compare on (a) ColoredMNIST~\citep{arjovsky2019invariant} and (b) AffNIST~\citep{affnist} the sprites learned with sprite-specific transformations (top rows) with the ones learned with shared transformations (bottom rows). Sharing the transformations among sprites encourages them to be more uniform, e.g., have similar (a) colors and (b) spatial location.}
    \label{fig:shared-tr-qual}
\end{figure}
Sharing transformations among sprites would intuitively put them in the same ``reference frame'' which would be beneficial for qualitative analysis. 
We visualize this effect in~\cref{fig:shared-tr-qual} on the ColoredMNIST and AffNIST datasets. When transformations are not shared, sprites have non-uniformed colors and positions, while they are much more consistent when transformations are shared. 
Although this qualitative property would be desirable, we found in the quantitative results reported in~\cref{tab:tr-shared} that sharing transformations significant deteriorates quantitative performances. \textbf{We thus keep sprite-specific transformations for each sprite in the rest of the analysis.}

%

\subsection[Decision and Training Criteria]{
\orangecircle{} Decision \Module{} and 
\yellowcircle{} Training Criteria
}
\label{subsec:res_dec}
%
%
Training criteria and decision \module{}s are closely related. Thus, we first analyze jointly decision \module{} and training criteria, then study the impact of regularizations.

\subsubsection{Reconstruction Loss and Decision}
\begin{table*}[ht]
    \centering
    \caption{\textbf{Results of different decision \module{}s (\orangecircle{}) and training criteria (\yellowcircle{}).} We experimented with the training criteria and decision \module{}s, along with Gumbel softmax. We report accuracy (\%) and standard deviation over 10 runs. We train all models and the baseline (second row)  until convergence, which might mean a different number of iterations for different models. \greencircle{}: MLP, \bluecircle{}: one-by-one, sprite-specific transformation.} 
    \tabcolsep=2pt
    \resizebox{\textwidth}{!}{
    \begin{tabular}{@{}l l c c c c c c c c c@{}}
    \toprule
    $\mathcal{L}_{\text{rec}}$ & $p_k$ & MNIST & ColoredMNIST & FashionMNIST & AffNIST & USPS & FRGC & SVHN & GTSRB-8 & Average \\
    \midrule
    \multirow{2}{*}{$\mathcal{L}_{\text{comp}}=\mathcal{L}_{0-1}$} & Min-Loss & 92.4±1.6 & \underline{71.0±2.1} & 58.3±0.6 & \underline{89.2±2.2} & 82.1±1.7 & 30.2±0.7 & \textbf{47.1±1.8} & 47.1±1.1 & 64.7 \\ 
    &\quad \textit{w/ reassignment} & \underline{96.5±0.5} & \textbf{92.0±2.0} & 59.6±0.7 & \textbf{97.3±0.0} & \textbf{88.4±2.9} & \textbf{41.1±0.6} & \underline{43.3±2.7} & 49.3±1.3  & 70.9 \\
    \midrule
    \multirow{2}{*}{$\mathcal{L}_{0-1}$} & weight prediction & 86.6±1.2 & 42.0±3.2 & 55.3±0.9 & 66.6±4.1 & 79.9±1.4 & 11.2±0.6 & 33.1±0.9 & \textbf{51.7±0.2} & 53.3 \\
    & linear mapping & 88.8±1.5 & 33.0±4.9 & 54.7±1.1 & 55.5±1.7 & 73.7±3.1 & 17.9±0.7 & 31.1±1.3 & \underline{51.6±0.4} &  50.8 \\
    \midrule
    \multirow{4}{*}{$\mathcal{L}_{\text{comp}}$} & weight prediction & 72.4±0.9 & 50.5±3.6 & 35.1±1.3 & 72.7±2.2 & 54.3±2.1 & \underline{40.2±0.9} & 19.7±0.3 & 38.2±1.0 & 47.9 \\
    & \quad\textit{w/ Gumbel softmax} & 93.2±1.6 & 47.7±4.5 & \underline{60.6±0.4} & 75.8±1.4 & 82.1±0.2 & 38.7±0.8 & 34.7±0.7 & 50.3±0.1 & 60.4 \\
    & linear mapping & 72.1±1.4 & 47.5±1.6 & 36.3±1.1 & 66.9±0.9 & 54.3±2.0 & 40.4±1.1 & 19.9±0.5 & 38.5±0.1 & 47.0 \\
    & \quad\textit{w/ Gumbel softmax} & \textbf{96.5±0.1} & 53.2±4.3 & \textbf{60.7±0.8} & 75.4±2.3 & \underline{82.2±0.4} & 39.5±1.3 & 33.9±0.5 & 50.0±0.2 & 61.4 \\
    \bottomrule
    \end{tabular}}
    \label{tab:proba-clustering}
\end{table*}
In~\cref{tab:proba-clustering}, we compare the reconstruction losses we introduced -- namely $\mathcal{L}_{0-1}$ defined in \cref{eq:loss-diff} and $\mathcal{L}_\text{comp}$ defined in \cref{eq:loss-sprite} -- alongside the different decision modules. \textit{Min-Loss} selection, for which both losses are the same, using a cluster re-assignment strategy~\citep{monnier2020dticlustering} (\cref{tab:proba-clustering} row 2) shows overall better performance than training the network to predict the sprite selection. This higher performance is largely due to the implicit regularization given by the empty cluster reassignment strategy proposed in ~\citep{monnier2020dticlustering} (\cref{tab:proba-clustering} rows 1 and 2).

Qualitatively, the main failure case of $\mathcal{L}_\text{comp}$ is to compose a layer from several sprites, as can be seen in~\cref{subfig:cri-qual-mnist} for MNIST, where a 9 digit is reconstructed using a circle and a loop, and in~\cref{subfig:cri-qual-frgc} for FRGC, where different sprites are combined to model lighting effects. As optimizing reconstruction by composition is not the targeted behavior for clustering, we experimented with replacing softmax activation with Gumbel softmax for this $\mathcal{L}_\text{comp}$, both with linear mapping and weight prediction (\cref{tab:proba-clustering} rows 6 and 8). This resulted in a significant improvement in performances of more than $10\%$ on average. While performances remains lower than with \textit{Min-Loss} selection with reassignment by almost $10\%$, this led to the best results with a predicted sprite selection, almost on par with \textit{Min-Loss} selection without re-assignment regularization. Learning sprite selection is appealing as it does not require to test all selection possibilities, {as in \textit{Min-Loss} selection, which will be prohibitively costly when using multiple layers. 
Because we obtained slightly better performances with linear mapping than classification weight prediction, and because it is conceptually simpler, \textbf{we use linear mapping for the rest of the paper}, and 
explore if its performance could be further improved using additional regularization losses. 
\begin{figure}[t]
    \centering
    \begin{subfigure}{0.45\columnwidth}
        \centering
        \includegraphics[width=\columnwidth,trim=0 0 0 3cm,clip]{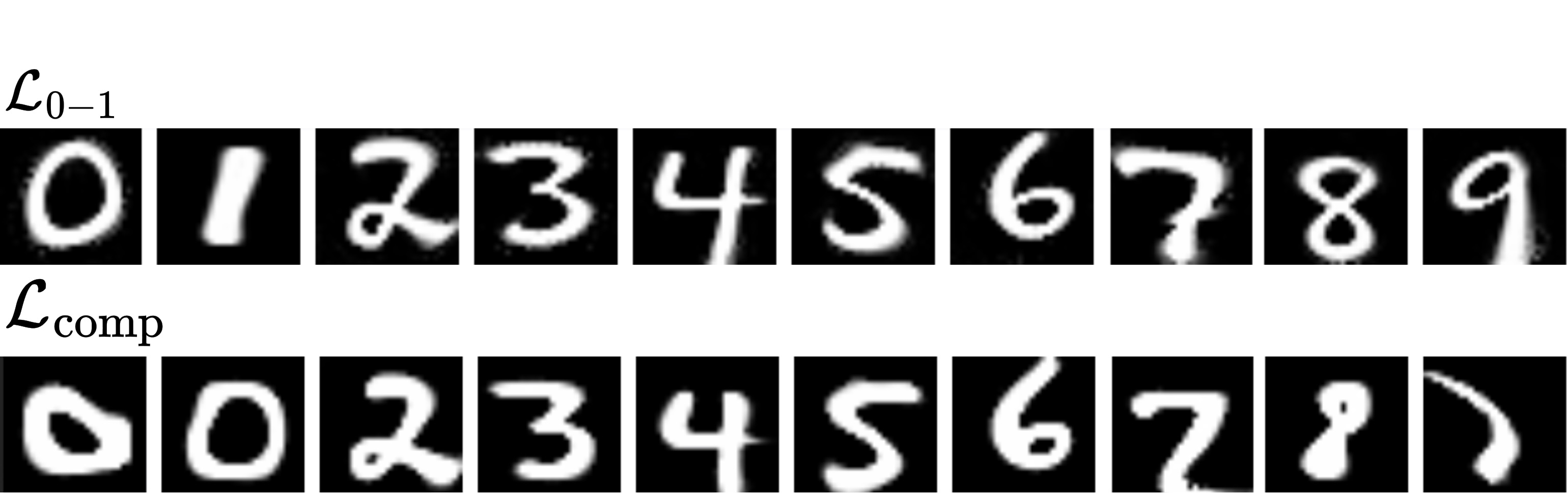}
        \caption{MNIST~\cite{lecun2010mnist}}
        \label{subfig:cri-qual-mnist}
    \end{subfigure}
    \hfill
    \begin{subfigure}{0.45\columnwidth}
        \centering
        \includegraphics[width=\columnwidth,trim=0 0 0 3cm,clip]{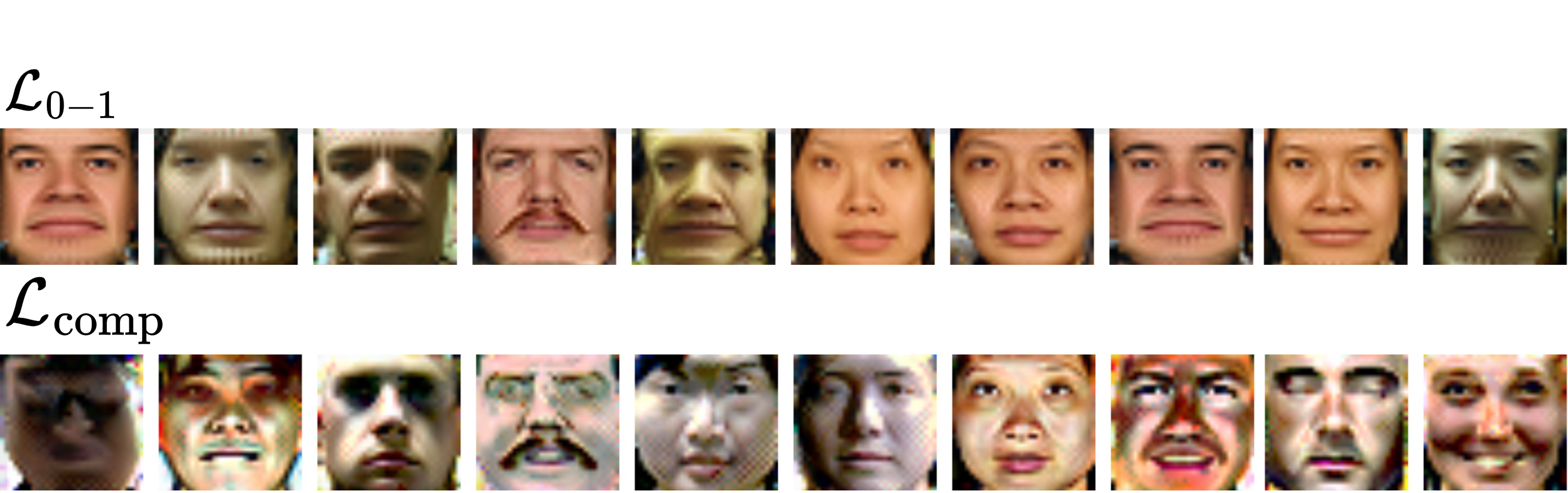}
        \caption{FRGC~\cite{phillips2005frgc}}
        \label{subfig:cri-qual-frgc}
    \end{subfigure}
    \caption{\yellowcircle{} \textbf{Qualitative results with different training criteria.} Compared with weighting the reconstruction loss for each sprite ($\mathcal{L}_{0-1}$, top rows), weighting transformed sprites and composing to reconstruct ($\mathcal{L}_\text{comp}$, bottom rows) results in (a) sprites representing parts of the objects instead of the object itself and (b) sprites focusing on the distinct characteristics of a subject and using composition to model shading effects.}
    \label{fig:criterion-qual}
\end{figure}
%


%
\begin{table*}[t]
    \centering
    \caption{\yellowcircle{} \textbf{Effect of regularization.} Experiments on regularization losses with two training criteria and Gumbel softmax. We report accuracy (\%) and standard deviation over 10 runs. \greencircle{}: MLP, \bluecircle{}: one-by-one, sprite-specific transformation, \orangecircle{}: linear mapping.}
    \tabcolsep=2pt
    \resizebox{\columnwidth}{!}{
    \begin{tabular}{@{}l l l c c c c c c c c c@{}}
    \toprule
    $\mathcal{L}_{\text{rec}}$ & $\mathcal{L}_{\text{freq}}$ & $\mathcal{L}_{\text{bin}}$ & MNIST
    & ColoredMNIST 
    & FashionMNIST 
    & AffNIST & USPS & FRGC & SVHN & GTSRB-8 & Average \\
    \midrule
    \multirow{2}{*}{$\mathcal{L}_{0-1}$} & - & - & 88.8±1.5 & 33.0±4.9 & 54.7±1.1 & 55.5±1.7 & 73.7±3.1 & 17.9±0.7 & 31.1±1.3 & 51.6±0.4 & 50.8 \\
    & \checkmark & - & \textbf{98.2±0.0} & 93.1±2.1 & 57.6±1.5 & \textbf{97.1±0.1} & \underline{82.8±0.1} & 41.1±0.6 & \textbf{38.0±2.0} & \textbf{57.5±0.2} & 70.7 \\
    \midrule
    \multirow{8}{*}{$\mathcal{L}_{\text{comp}}$} & \multicolumn{2}{l}{\textit{w/ softmax}} & & & & & & & & & \\
    & - & - & 72.1±1.4 & 47.5±1.6 & 36.3±1.1 & 66.9±0.9 & 54.3±2.0 & 40.4±1.1 & 19.9±0.5 & 38.5±0.1 & 47.0 \\
    & \checkmark & - & 78.5±1.6 & 48.4±3.1 & 40.6±1.1 & 75.5±0.0 & 54.3±2.0 & 42.6±1.1 & 23.4±0.9 & 38.8±0.3 & 50.3 \\
    & \checkmark & \checkmark & 95.3±0.5 & 81.7±3.5 &  \textbf{62.0±1.5} & 83.1±0.0 & 63.1±2.2 & 42.6±1.1 & 34.4±0.5 & \underline{54.5±2.1} & 64.6 \\
    & \multicolumn{2}{l}{\textit{w/ Gumbel softmax}} & & & & & & & & & \\
    & - & - & 96.5±0.1 & 53.2±4.3 & \underline{60.7±0.8} & 75.4±2.3 & 82.2±0.4 & 39.5±1.3 & 33.9±0.5 & 50.0±0.2 & 61.4 \\
    & \checkmark & - & \underline{96.7±0.0} & \underline{95.9±0.1} & \underline{60.7±0.8} & \underline{94.1±1.9} & 82.2±0.4 & \textbf{44.8±0.8} & 35.3±0.4 & 53.2±1.2 & 70.4 \\
    & \checkmark & \checkmark & \underline{96.7±0.0} & \textbf{96.0±0.1} & \underline{60.7±0.8} & \underline{94.1±1.9} & \textbf{85.3±1.1} & \textbf{44.8±0.8} & \underline{37.6±0.3} & 53.2±1.2 & 71.1 \\
    \bottomrule
    \end{tabular}}
    \label{tab:proba-regularization}
\end{table*}
\subsubsection{Regularization Losses}
\label{subsubsec:reg} We report in~\cref{tab:proba-regularization} the results obtained with using $\mathcal{L}_\text{freq}$ (\cref{eq:loss_reg_freq}) and $\mathcal{L}_{\text{bin}}$. Note that $\mathcal{L}_{\text{bin}}$ is designed to overcome the composition issue associated to $\mathcal{L}_\text{comp}$, we do not test it with $\mathcal{L}_\text{0-1}$, and $\mathcal{L}_{\text{empty}}$ does not make sense for clustering, where no empty sprite is used. 
We selected the regularization loss weights through a grid search for each dataset to optimize performance.

Using $\mathcal{L}_\text{freq}$ (\cref{eq:loss_reg_freq}) as a regularization significantly increases performance. Both when using $\mathcal{L}_{0-1}$ and $\mathcal{L}_\text{comp}$ losses coupled with Gumbel softmax, this leads to results on par with \textit{Min-Loss} selection with reassignment (\cref{tab:proba-regularization}). This again validates our claim that the superior performance of \textit{Min-Loss} selection is largely due to the implicit regularization of the reassignment strategy.

To improve results obtained with $\mathcal{L}_\text{comp}$ we evaluated using $\mathcal{L}_{\text{bin}}$ (\cref{eq:loss_reg_bin}) to encourage binary selection, similar to~\cite{smirnov2021marionette}. $\mathcal{L}_{\text{bin}}$ significantly improves the results with normal softmax while remaining worse than the best approaches, and provides a small improvement when using Gumbel softmax which already encourages binary selection. 
{\bf We thus propose using $\mathcal{L}_\text{comp}$ with Gumbel softmax, and $\mathcal{L}_\text{freq}$ and $\mathcal{L}_{\text{bin}}$ regularizations.}



\subsection{Comparison with State-of-the-art}
\label{sec:clustering_comp}
Given the analysis of the previous sections, we use the following design choices, summarized in~\cref{tab:module-setup}, for clustering: using an MLP-based sprite generation module, with sprite-specific transformations, learned one-by-one in a curriculum fashion, a linear decision module with Gumbel softmax, a composite reconstruction loss, and frequency and binning regularization. We compare the performance of this setting with a single opaque layer (Ours-C 1 layer) to a variety of competing clustering methods in~\cref{tab:clustering-comp}. For SVHN and GTSRB-8, we also report our approach using a background model (Ours-C 2 layers). Note that most approaches rely on learning and clustering features, which limit the results' interpretability, and that many leverage specific data-augmentation or representations, such as Gabor filters, which are strong priors and simplify the task. Our results are competitive with state-of-the-art on class-aware metrics, while predicting cluster selection, and learning an explicit cluster prototype and image-specific transformation. While it often performs slightly worse than DTI-Clustering, our setting does not rely on comparing each possible reconstruction to the target to assign clusters, but instead directly learns and predicts cluster selection. Thus, as shown in the next session, our approach can be directly extended into an efficient multi-layer image decomposition model.

\begin{table*}[t]
    \centering
    \caption{\textbf{Comparisons on clustering.} We compare our results with methods that cluster over features as well as pixels. We report accuracy (\%) and standard deviation for our method over 10 runs.}
    \tabcolsep=2pt
    \resizebox{\columnwidth}{!}{
    \begin{tabular}{@{}llcccccccc@{}}
    \toprule
    Method & \# runs &  MNIST & ColoredMNIST & FashionMNIST & AffNIST & USPS & FRGC & SVHN & GTSRB-8 \\
    \midrule
    \multicolumn{10}{l}{\textit{Clustering on learned features}} \\
    JULE~\cite{yang2016joint} & 3 & 96.4 & - & \underline{56.3} & - & \underline{95.0} & \underline{46.1} & - & - \\
    DEPICT~\cite{dizaji2017deep} & 5 & 96.5 & - & 39.2 & - & \textbf{96.4} & \textbf{47.0} & - & - \\
    DSCDAN~\cite{yang2019deep} & 10 & 97.8 & - & \textbf{66.2} & - & 86.9 & - & - & - \\
    \multicolumn{10}{l}{\textit{+ with data augmentation and/or ad-hoc data representation}} \\
    SpectralNet~\cite{shaham2018spectralnet} & 5 & 97.1 & - & - & - & - & - & - & - \\ 
    IMSAT~\cite{hu2017learning} & 12 (5) & 98.4 & (10.6) & - & (18.2) & - & - & \textbf{57.3} & 26.9  \\
    ADC~\cite{haeusser2019adc} & 20 & \textbf{98.7} & - & - & - & - & 43.7 & 38.6 & - \\
    SCAE~\cite{kosiorek2019stacked} & 5 & \textbf{98.7} & - & - & - & - & - & \underline{55.3} & - \\
    IIC~\cite{ji2019invariant} & 5 & 98.4 & 10.6 & - & \textbf{57.6} & - & - & - & - \\
    SCAN~\cite{van2020scan} & 5 & - & - & - & - & - & - & 54.2 & \textbf{90.4} \\ 
    \midrule
    \multicolumn{10}{l}{\textit{Clustering on pixels}} \\
    K-means & 10 & 54.8 & - & 54.1 & - & 65.3 & 22.7 & 12.2 & - \\
    DTI-Clustering~\cite{monnier2020dticlustering} & 10 & \textbf{97.3} & \textbf{96.8} & \textbf{61.2} & \textbf{95.5}  & \textbf{86.4} & \underline{39.6} & 44.5 & - \\
    \textbf{Ours-C} 1 layer & 10 & \underline{96.7±0.0} & \underline{96.0±0.1} & \underline{60.7±0.8} & \underline{94.1±1.9} & \underline{85.3±1.1} & \textbf{44.8±0.8} & 37.6±0.3 & 53.2±1.2 \\
    \multicolumn{10}{l}{\textit{+ multi-layer}} \\
    DTI-Sprites~\cite{monnier2021unsupervised} & 10 & - & - & - & - & - & - & \textbf{63.1} & \textbf{89.9} \\
    \textbf{Ours-C} 2 layers & 10 & - & - & - & - & - & - & \underline{52.4±0.5} & \underline{80.9±1.8} \\
    \bottomrule
    \end{tabular}}
    \label{tab:clustering-comp}
\end{table*}

%% file: Sections/5-MultiObject.tex
\section{Analysis for Multi-layer Image Decomposition}
\label{sec:analysis_decomp}
In this section, we explore how our analysis of sprite-based image models for clustering can be leveraged for multi-layer image decomposition. In the following, we first summarize our experimental setting, including datasets and metrics, and then discuss our results.

\subsection{Experimental Setup}
\subsubsection{Datasets} We present results on the Tetrominoes~\citep{greff2019multiobject} -- images of 3 non-overlapping colored 2D blocks sampled among 19 unique ones, on black background --, the Multi-dSprites~\citep{kabra2019multi} -- images of up to 5 possibly overlapping colored 2D objects of different sizes sampled among 3 unique ones, on gray background --, and the CLEVR6 and CLEVR~\citep{johnson2017clevr} datasets -- images of respectively up to 6 and 10 possibly overlapping colored 3D objects of different sizes sampled among 6 unique ones, on a simple background. More details are given in the~\cref{sec:supmat_data}. Note that all of these datasets are synthetic and relatively simple, but they are the main ones used in the literature for our task.

\subsubsection{Training Details} Details of the transformations for the foreground and background are given in the Appendix~\cref{tab:tr-selection}. Our empirical results across four distinct datasets show that the regularization weight $\lambda_\text{empty}$ is crucial for multi-layer object decomposition, and higher scene complexity generally demands higher $\lambda_\text{empty}$. Details of training setup and hyperparameters for each dataset are provided in the~\cref{app:training}.

\subsubsection{Metrics}
For our analysis in~\cref{tab:obj-disc}, we reported two class-aware metrics, mean accuracy (mAcc) and average mean Intersection-over-Union (avg-mIoU). We use Hungarian matching to align the predicted and ground-truth classes. Mean accuracy measures the proportion of correctly-predicted pixels in classification. The average mean IoU computes the IoU, a segmentation accuracy metric, class-wise and averages it across all classes, including background, to reflect class awareness. 

To give results comparable with the metrics most frequently reported in the literature, we also report in~\cref{tab:inst-seg} instance mean IoU (mIoU) and the Adjusted Rand Index computed only for the foreground (ARI-FG). 
Instance mean IoU measures the segmentation accuracy of predicted instances without considering the class of the prediction, but takes the background into account. ARI-FG  evaluates how well pixels' instance assignments align with the ground truth while ignoring the background. These two metrics are adopted by the literature because most existing methods focus solely on predicting instance segmentation without providing the corresponding class labels. For the few approaches that additionally provide class predictions, we also report mean accuracy (mAcc) and mean IoU averaged over classes (avg-mIoU). 

\subsection{Results}
\label{subsec:res_multi}
\subsubsection{Regularizations}

\begin{table*}[t!]
    \centering
\caption{\textbf{Results for multi-object semantic discovery.} (\greencircle{}) Sprite generation, (\orangecircle{}) decision and activation function, (\yellowcircle{}) training criterion and regularization. Mean accuracy (mAcc) and average mean IoU (avg-mIoU) over classes, with standard error over 3 runs. ($\dagger$): longer training, except \citet{monnier2021unsupervised} on CLEVR (in \textit{italic}) obtained with the training schedule in the paper. \textcolor{gray}{Gray} entries denote single-run results where initial performance indicated lower performance than the baseline.
    }
    \tabcolsep=2pt
    \resizebox{\columnwidth}{!}{
    \begin{tabular}{@{}llllllcc?cc?cc?cc@{}}
    \toprule
    Method & \greencircle{} & \orangecircle{} & \orangecircle{} & \yellowcircle{} & \yellowcircle{} & \multicolumn{2}{c}{Tetrominoes} & \multicolumn{2}{c}{Multi-dSprites} & \multicolumn{2}{c}{CLEVR6} & \multicolumn{2}{c}{CLEVR} \\
    \midrule
   & & & & & & \underline{mAcc} & \underline{avg-mIoU} & \underline{mAcc} & \underline{avg-mIoU} & \underline{mAcc} & \underline{avg-mIoU} & \underline{mAcc} & \underline{avg-mIoU} \\
    DTI-Sprites~\cite{monnier2021unsupervised}$^\dagger$ & Pixels &Min-Loss & S & $\mathcal{L}_{0-1 =\text{comp}}$ & $\mathcal{L}_\text{empty}$ & \textbf{99.5±0.2} & \textbf{99.2±0.3} & \textbf{91.3±0.9} & \textbf{84.0±1.4} & \textbf{79.3±2.7} & \textbf{64.2±3.1} & \textit{69.8±5.0} &	\textbf{\textit{55.7±4.3}}\\
    \midrule 
    \multirow{7}{*}{\textbf{Ours-D}} & \multirow{7}{*}{MLP} & \multirow{7}{*}{\parbox[t]{1.5cm}{\raggedright linear mapping\par}} & \multirow{7}{*}{GS} & \multirow{7}{*}{$\mathcal{L}_{\text{comp}}$} & - & 74.3±1.4 & 64.4±1.9 & 65.6±0.1 & 54.4±0.1 & 66.8±0.4 & 49.6±0.8 & 59.3±7.4 & 43.0±5.6 \\
    & & & & & $\tau$ & 92.7±3.9 & 89.2±5.5 & \color{gray} 65.2 & \color{gray} 53.8 & \color{gray} 56.3 & \color{gray} 39.6 & \color{gray} 55.3 & \color{gray} 42.1 \\
    & & & & & + $\mathcal{L}_\text{freq}$ & \underline{93.9±0.4} & \underline{89.9±0.5} & - & - & - & - & - & - \\
    & & & & & $\mathcal{L}_\text{empty}$ & - & - & 65.9±0.5 & 54.7±0.4 & \underline{74.7±1.4} & \underline{57.8±1.5} & \textbf{70.6±0.1}	& \underline{55.3±0.1} \\ 
    & & & & & + $\mathcal{L}_\text{freq}$ & - & - & \underline{66.0±1.0} & \underline{54.7±0.9} & 72.2±1.2 & 54.8±1.3 & \underline{70.5±1.2} & 53.9±0.5 \\
    & & & & & + $\mathcal{L}_\text{bin}$ & - & - & \color{gray} 65.4 & \color{gray} 54.2 & \color{gray} 65.0 & \color{gray} 46.5 & \color{gray} 68.0 & \color{gray} 53.1 \\
    \bottomrule
    \end{tabular}}
    \label{tab:obj-disc}
\end{table*}
In~\cref{tab:obj-disc}, we analyze the impact of different regularizations on the performance of our approach and compare it to DTI-Sprites~\citep{monnier2021unsupervised}. 
Indeed, the regularization needs are different from the ones in clustering. In particular, $\mathcal{L}_\text{bin}$, which we adopted for clustering, prevents multiple sprites from the same layer from being combined to reconstruct different parts of the same object, but it does not prevent the same effect with sprites from different layers. Thus, in addition to $\mathcal{L}_\text{bin}$ and $\mathcal{L}_\text{freq}$, we experimented with $\mathcal{L}_\text{empty}$ (\cref{eq:empty}) which favors the use of an empty sprite, \ie a sprite with a completely transparent alpha mask.
Because we observed early-stage high-confidence class predictions during training, which is likely detrimental to learning, we also explored the impact of learning the Gumbel softmax temperature parameter, $\tau$, which could mitigate this effect. 

For Tetrominoes, where the number of objects is constant and equal to the number of layers, $\mathcal{L}_\text{empty}$ and $\mathcal{L}_\text{bin}$ make little sense, and we only explore learning the Gumbel softmax temperature $\tau$ and $\mathcal{L}_\text{freq}$, while we explore all regularizations for the other datasets. 

Learning the Gumbel softmax temperature $\tau$ gives a huge boost to the results on Tetrominoes. One of the three runs actually matches the almost perfect performance of DTI-Sprites, emphasizing the additional complexity of learning the class prediction. Adding $\mathcal{L}_\text{freq}$ to learning $\tau$ further improves the average on Tetrominoes, but they remain below the almost perfect results of DTI-Sprites, without any run matching it. 
In contrast, for Multi-dSprites, CLEVR6, and CLEVR, learning $\tau$ leads to the worst results.

For Multi-dSprites, CLEVR6, and CLEVR, we thus discarded learning $\tau$ and instead experimented with $\mathcal{L}_\text{empty}$, trying to better estimate the number of objects, which is the main challenge for our approach without regularization on CLEVR6 and CLEVR. On Multi-dSprites~\citep{greff2019multiobject}, which includes 3 distinct objects (square, ellipsoid, and heart), our method is still clearly outperformed by DTI-Sprites~\citep{monnier2021unsupervised}, due to the fact that our model fails to discover a distinct representation of the heart shape, instead reconstructing it as a composition of two rotated ellipsoids. 
On CLEVR6 and CLEVR, our performance with $\mathcal{L}_\text{empty}$ is on par with DTI-Sprites, and the results are qualitatively similar. Further adding $\mathcal{L}_\text{freq}$ does not significantly change these results, and adding $\mathcal{L}_\text{bin}$ degrades them. The main difference between our approach and DTI-Sprites on this more challenging CLEVR dataset is the higher scalability of our approach, which we discuss in the next section. 

\subsubsection{Complexity}
\begin{wrapfigure}{r}{0.6\columnwidth}
    \centering
        \includegraphics[width=0.4\columnwidth]{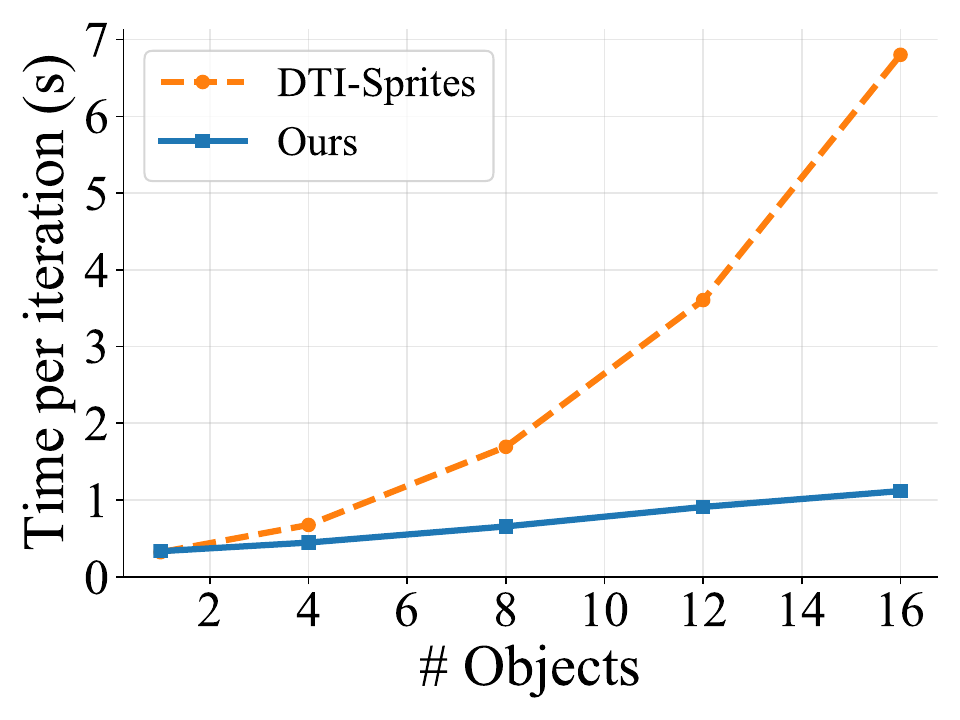}
        \caption{\textbf{Complexity.} The time per iteration of our approach scales linearly with the number of object layers, while that of the only other method with comparable results, DTI-Sprites~\citep{monnier2021unsupervised}, scales exponentially.}
    \label{fig:time}
\end{wrapfigure}

The major advantage of our model with respect to DTI-Sprites is that it learns to predict which sprite to select for which layer, rather than iteratively trying many sprite selections and combinations, which results in a significant improvement in time complexity, as shown in~\cref{fig:time} \edit{on the CLEVR dataset}. 

The time complexity of our model scales linearly with the maximum number of objects in a scene, while the DTI-Sprites scales exponentially. {Note that due to this exponential time complexity, we trained DTI-Sprites in~\cref{tab:obj-disc} and~\cref{tab:inst-seg} with the training schedule in the original paper, \ie 351,900 iterations, while we could train our model until convergence, for 703,800 iterations, in less time.}

\subsubsection{Comparisons}
In~\ref{tab:inst-seg}, we compare our results with the state-of-the-art on the CLEVR dataset. AST-Seg-B3-CT~\citep{sauvalle2023unsupervised} clearly dominates in terms of mIoU and ARI-FG, but our results are on par with most baselines for these metrics, which focus on instances and do not consider the class prediction.  We thus also compared class aware metrics for methods from which can be extracted.
For MarioNette~\citep{smirnov2021marionette}, we match learned sprites in the dictionary to classes in a many-to-one manner with Hungarian matching. 
Because it is the best-performing instance segmentation method, we also applied K-means with $K=6$ to the object features ($z_\text{what}$) of AST-Seg-B3-CT~\citep{sauvalle2023unsupervised} and clustered them, leading to a class-aware adaptation of this method.
The only method that achieves similar results to ours for class-aware metrics is DTI-Sprites~\citep{monnier2021unsupervised}, while the other two baselines lag far behind. This emphasizes another advantage of our approach.

\subsubsection{Qualitative Results}
\begin{figure*}[t!]
    \centering
    \includegraphics[width=\textwidth]{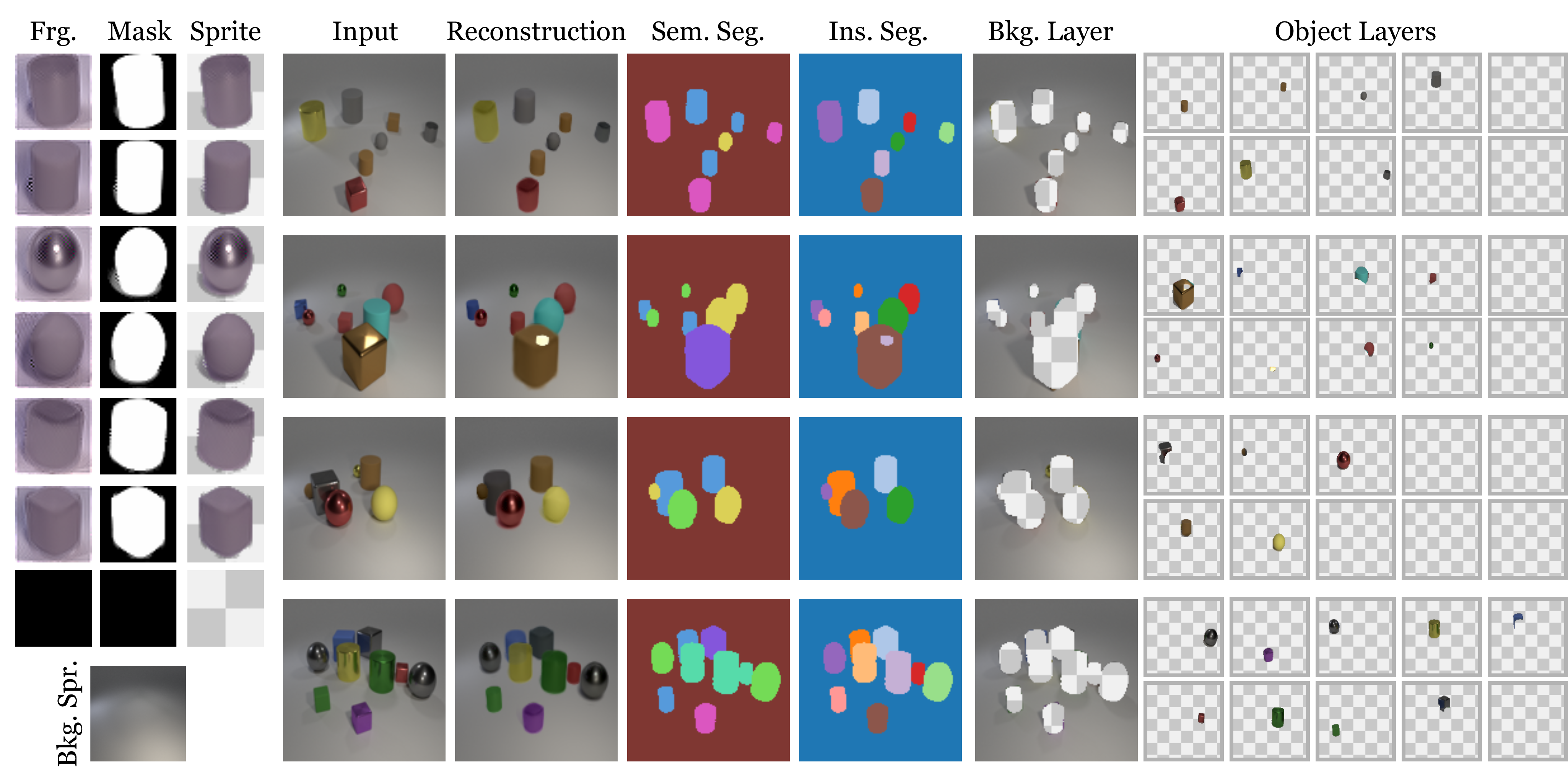}
    \caption{\textbf{Qualitative results for multi-object discovery on CLEVR~\citep{johnson2017clevr}.} The three left columns show the sprites' appearances (Frg.), masks, and combination (Sprite), including the empty sprite, and the background. The other columns show for four different examples, the input image, its reconstruction, semantic segmentation (Sem. Seg.), instance segmentation (Ins. Seg.), background (Bkg. Layer), and the different transformed sprites (Object Layers).}
    \label{fig:qual-discovery}
\end{figure*}

Qualitative examples of our decomposition with a large number of objects from CLEVR~\citep{johnson2017clevr} are presented in~\cref{fig:qual-discovery}. They show that our model is able to recover both accurate instances and semantic segmentation with a large number of objects.

\begin{table}[t!]
    \centering
    \caption{\textbf{Comparisons for instance segmentation} with standard deviation over 3 runs. Sources for $\dagger$ (excluding~\citet{monnier2021unsupervised}):~\cite{karazija2021clevrtex} and~\cite{sauvalle2023unsupervised}.
    }
    \tabcolsep=3pt
    \resizebox{0.75\textwidth}{!}{
    \begin{tabular}{@{}lccccc@{}}
    \toprule
    Method & class-aware & \multicolumn{4}{c}{CLEVR} \\
    \midrule
    & & \underline{mIoU$^\dagger$} & \underline{ARI-FG$^\dagger$} & \underline{mAcc} & \underline{avg-mIoU} \\
    MONet~\cite{burgess2019monet} && 30.7±14.9 & 54.5±11.4 & - & - \\
    IODINE~\cite{greff2019multiobject} && 45.1±17.9 &\cellcolor{g4}{93.8±0.8} & - & - \\
    SPAIR~\cite{crawford2019spatially} && \cellcolor{g6}{66.0±4.0} & \cellcolor{g8}{77.1±1.9} & - & - \\
    GNM~\cite{jiang2020generative} &&\cellcolor{g7}{59.9±3.7} & \cellcolor{g10}{65.1±4.2} & - & -  \\
    Slot Attention~\cite{locatello2020objectcentric} &&36.6±24.8  & \cellcolor{g3}{95.9±2.4} & - & - \\
    eMORL~\cite{emami2021efficient} && 50.2±22.6 & \cellcolor{g5}{93.3±3.2} & - & - \\
    Genesis-V2~\cite{engelcke2021genesis} && 9.5±0.6 & 57.9±20.4 & - & - \\
    MarioNette~\cite{smirnov2021marionette} &\checkmark & \cellcolor{g3}{72.1±0.6} & 56.8±0.4 & \cellcolor{g9}{16.1±0.2} & \cellcolor{g10}{7.3±0.4} \\
    AST-Seg-B3-CT~\cite{sauvalle2023unsupervised} & & \cellcolor{g2}{90.3±0.2} & \cellcolor{g1}{98.3±0.1} & \cellcolor{g7}{20.8±1.2}&\cellcolor{g9}{12.1±0.2} \\

    \midrule
    DTI-Sprites~\cite{monnier2021unsupervised} & \checkmark & \cellcolor{g8} 54.5±1.2 & \cellcolor{g6}{93.2±2.0} & \cellcolor{g4}{69.8±4.5} & \cellcolor{g4}{55.7±6.0} \\
    \textbf{Ours-D} &\checkmark & \cellcolor{g8}{53.8±0.3} & \cellcolor{g3}{95.1±0.5} & \cellcolor{g2}{70.6±0.2} & \cellcolor{g4}{55.3±0.2} \\ 
   
    \bottomrule
    \end{tabular}}
    \label{tab:inst-seg}
\end{table}

%% file: Sections/6-Conclusion.tex
\section{Conclusion}
In this work, we introduced a unified formalization for sprite-based models, specifying their relationships, and unifying approaches for clustering and multi-layer image decomposition. This analysis clarifies the design space of methods in the literature and enables its exploration on the clustering task, which is less computationally intensive and uses more diverse, realistic datasets. This yields in turn an approach to image decomposition that learns to predict sprite selection, to avoid an exponential complexity in the number of objects per image, and maintains strong performance. 

Our study offers several {key insights} for layer-based image decomposition: \textbf{(i)} learning sprite representations via an MLP yields slightly better reconstruction, but more importantly significantly accelerates training; \textbf{(ii)} sharing transformation parameters across sprites acts as a structural regularizer, encouraging uniformity in color and spatial alignment, but can slightly hurt performances; 
\textbf{(iii)} opposite to the exponential cost of min-loss optimization, predicting sprite probabilities and using a loss on composite sprites scales linearly with the number of objects per image, enabling to model a larger number of objects per image, but requires explicit regularization and performs slightly worse in simple cases. Moreover, our analysis emphasizes that while the standard CLEVR benchmarks seem saturated in terms of the commonly reported class-agnostic metrics, it is actually far from being the case for class-aware metrics, where our method provides state-of-the-art but imperfect results, encouraging more works in this direction.

One of the main limitations of the approach we propose is the necessity of hyperparameter selection. Although we follow standard protocols in the literature for evaluation and show some robustness to hyperparameter selection, our reliance on ground-truth validation labels to perform a grid search for regularization hyperparameters departs from a fully unsupervised deployment setting where labels are unavailable.

%% file: Sections/7-Appendix.tex
\subsection{Dataset Descriptions}
\label{sec:supmat_data}

\paragraph{MNIST~\citep{lecun2010mnist}} MNIST is a widely used dataset of handwritten grayscale digits, containing 60,000 training images and 10,000 testing images.

\paragraph{ColoredMNIST~\citep{arjovsky2019invariant}} Colored MNIST is built from the MNIST dataset by randomly adding color to the foreground and background, resulting in a collection of 70,000 images. Each digit image is transformed into a 3-channel representation, offering a more complex dataset.

\paragraph{FashionMNIST~\citep{xiao2017fashion}} FashionMNIST is designed as an alternative to MNIST, consisting of 60,000 fashion item images for training and 10,000 for testing. These images are grayscale and categorized into 10 classes.

\paragraph{AffNIST~\citep{affnist}} Derived from MNIST, the AffNIST dataset enriches the original dataset by applying affine transformations to its digits. We employ the test split of 10,000 images to assess algorithm robustness against various transformations.

\paragraph{USPS~\citep{hull1994usps}} The United States Postal Service (USPS) dataset includes handwritten grayscale digit images of envelopes, containing 7,291 training samples and 2,007 testing samples.

\paragraph{FRGC~\citep{phillips2005frgc}} The Face Recognition Grand Challenge (FRGC) dataset is a collection of face images in RGB space, which contains over 50,000 images of various individuals captured under different poses, expressions, and lighting conditions.

\paragraph{SVHN~\citep{netzer2011reading}} The Street View House Numbers (SVHN) dataset includes more than 600,000 RGB images of house numbers captured from Google Street View. It is intended for digit recognition tasks and offers more challenging variations in terms of font styles, sizes, and cluttered backgrounds compared to MNIST.

\paragraph{GTSRB-8~\citep{stallkamp2012man}} The German Traffic Sign Recognition Benchmark (GTSRB) dataset subset (GTSRB-8) focuses on eight common traffic sign classes and contains more than 25,000 images for training and testing.

\paragraph{Tetrominoes~\citep{greff2019multiobject}} Tetrominoes contains around 60,000 images with size 35x35 featuring 3 Tetris-like shapes with different color and position from 19 unique shapes. Each image has a black background, and shapes do not occlude each other. 

\paragraph{Multi-dSprites~\citep{kabra2019multi}} Multi-dSprites contains around 60,000 images with multiple oval, heart, or square-shaped objects with a uniform background. Each object has different scale, color, and position, and the maximum number of objects in an image is 5.

\paragraph{CLEVR~\citep{johnson2017clevr}} CLEVR dataset contains 6 unique objects with varying scale, color, and position on a uniform background. Although released for visual reasoning tasks, it is commonly used in object discovery. We reported results in 2 versions of CLEVR: CLEVR6 and CLEVR where the maximum numbers of objects in an image are 6 and 10, respectively. CLEVR6 contains around 35,000 and CLEVR contains around 100,000 images.

\subsection{Training Details}
\label{app:training}
We adopt the training setup of~\citet{monnier2020dticlustering} for clustering and~\citet{monnier2021unsupervised} for multi-object semantic discovery as our baseline. Hyperparameters are provided in~\cref{tab:cluster-setup,tab:disc-setup}. For~\cref{tab:obj-disc}, we report the mean and standard error of 3 runs. Due to its computational complexity, we adopt the training schedule reported for CLEVR6 in~\citet{monnier2021unsupervised} to CLEVR for DTI-Sprites (\textit{italic} in~\cref{tab:obj-disc}). To be comparable with the literature~\citep{karazija2021clevrtex}, we reported the mean and standard deviation of 3 runs for~\cref{tab:inst-seg}. The results for DTI-Sprites and our variation are reported over the whole dataset.
 \begin{table*}
    \centering
    \caption{\textbf{Training setup and hyperparameters for clustering.}}
    \tabcolsep=1pt
    \resizebox{\columnwidth}{!}{
    \begin{tabular}{@{}lcccccccc@{}}
    \toprule
    Dataset & MNIST & ColoredMNIST & FashionMNIST & AffNIST & USPS & FRGC & SVHN & GTSRB-8 \\
    \midrule
    \rowcolor{lightergray} \textit{Model \& Data} \\
    \# sprites	& 10 & 10 & 10 & 10 & 10 & 20 & 10 & 8 \\
    sprite tr.	& id, aff, mor, tps & id, color, aff, tps & id, color, aff, tps & id, aff, mor, tps & id, color, aff, tps & id, color, aff, tps  & id, color, proj & id, color, proj \\
    sprite tr. curr. & 10, 30, 40 & 10, 30, 60 & 10, 30, 50 & 10, 40, 50 & 120, 240, 400 & 100, 400, 800 & 16, 144 & 160, 1440\\
    \rowcolor{lightergray} \textit{Training} \\
    batch size & 128 & 128 & 128 & 128 & 128 & 128 & 128 & 128 \\
    learning rate & 1e-3 & 1e-3 & 1e-3 & 1e-4 & 1e-3 & 1e-3 & 1e-3 & 1e-3 \\
    weight decay & 1e-6 & 1e-6 & 1e-6 & 1e-6 & 1e-6 & 1e-6 & 1e-6 & 1e-6 \\
    lr. step & 70 & 90 & 70 & 74 & 500 & 1300 & 240 & 2400 \\
    \# epochs & 80 & 100 & 80 & 90 & 640 & 1400 & 264 & 2640 \\
    $\lambda_\text{freq}$ & 0.01 & 0.1 & 0 & 0.01 & 0 & 0.01 & 0.01 & 0.1 \\
    $\lambda_\text{bin}$ & 0 & 0.001 & 0 & 0 & 0.01 & 0 & 0.001 & 0 \\
    \bottomrule
    \end{tabular}}
    \label{tab:cluster-setup}
\end{table*}

\begin{table*}
    \centering
    \caption{\textbf{Training setup and hyperparameters for multi-object decomposition.}}
    \tabcolsep=2pt
    \begin{tabular}{@{}lcccc@{}}
    \toprule
    Dataset & Tetrominoes & Multi-dSprites & CLEVR6 & CLEVR \\
    \midrule
    \cellcolor{lightergray} \textit{Model \& Data} &&&&\\
    \# sprites	& 19 & 3 & 6 & 6 \\
    \# bkg & 0 & 1 & 1 & 1 \\
    \# objects & 3 & 5 & 6 & 10 \\
    \# channels	& 3 & 3 & 3 & 3 \\
    frg., bkg., mask curr. & 600, 0, 1 & 0, 0, 20 & 0, 0, 80 & 0, 0, 80 \\
    sprite/layer init. & cons, cons, gauss. & cons, cons, gauss. & cons, mean, gauss. & cons, mean, gauss. \\
    init. values & 0.9, 0.9, 0. & 0.9, 0.5, 0. & 0.9, 0., 0. & 0.9, 0., 0. \\
    gauss. std. & 5 & 7 & 10 & 10 \\
    sprite tr. & id & id, scale+rot. & id, proj. & id, proj. \\
    bkg. tr. & - & color & color & color \\
    layer tr. & color, scale+affine & color, scale+affine & color, scale+affine & color, scale+affine \\
    sprite tr. curr. &  - & 40 & 300 & 300 \\
    sprite size	& 24, 24 & 28, 28 & 40, 40 & 40, 40 \\
    image size	& 35, 35 & 35, 35 & 128, 128 & 128, 128 \\
    occlusion & - & \checkmark & \checkmark & \checkmark \\
    \cellcolor{lightergray} \textit{Training} &&&& \\
    avg. pool & 1, 1 & 1, 1 & 1, 1 & 1, 1 \\
    batch size & 32 & 32 & 32 & 32 \\
    learning rate & 1e-4 & 1e-4 & 1e-4 & 1e-4 \\
    lr. step & 1000, 1200 & 500, 1000 & 500,800 & 500, 800 \\
    \# epochs & 1220 & 1020 & 900 & 900 \\
    $\lambda_\text{freq}$ & 1e-3 & 0 & 0 & 0 \\
    $\lambda_\text{bin}$ & 1e-4 & 0 & 0 & 0 \\
    $\lambda_\text{empty}$ & - & 1e-4 & 1e-3 & 1e-2 \\
\bottomrule
    \end{tabular}
    \label{tab:disc-setup}
\end{table*}

\subsubsection{Transformation Module}
We follow the transformation setup and order in~\cref{tab:tr-selection} according to~\cite{monnier2020dticlustering, monnier2021unsupervised}.~\cref{tab:tr-selection} demonstrates three levels of transformations, applied to the sprites, the background and the layers.
%
\begin{table}[ht]
    \centering
    \tabcolsep=3pt
    \caption{\bluecircle{} \textbf{Transformation setups of datasets.} Transformations are selected and ordered depending on the characteristics of each dataset. Transformations for \hllightgray{background} and \hlgray{layers} are highlighted.}
    \resizebox{0.7\textwidth}{!}{
    \begin{tabular}{@{}l c c c c c c c c@{}}
    \toprule
        Dataset & id. & \color{blue} color & \color{blue} affine & \color{red} morpho. & \color{red} tps & \color{red} proj. & scale+rot. & scale+affine \\
        \midrule
        MNIST & 1 & & 2 & 3 & 4 & \\
        ColoredMNIST & 1 & 2 & 3 & & 4 & \\
        FashionMNIST & 1 & 2 & 3 & & 4 & \\
        affNIST & 1 & & 2 & 3 & 4 & \\
        USPS & 1 & 2 & 3 & & 4 & \\
        FRGC & 1 & 2 & 3 & & 4 & \\
        SVHN & 1 & 2 & & & & 3\\
        GTSRB-8 & 1 & 2 & & & & 3 \\
        \midrule
        Tetrominoes & 1/\hllightgray{1} & \hlgray{1} & & & & & & \hlgray{2} \\
        Multi-dSprites & 1 & \hllightgray{1}/\hlgray{1} & & & & & 2 & \hlgray{2} \\
        CLEVR(6) & 1 & \hllightgray{1}/\hlgray{1} & & & & & & \hlgray{2} \\
    \bottomrule
    \end{tabular}}
    \label{tab:tr-selection}
\end{table}

\subsubsection{Analysis on Regularization Hyperparameter Tuning}
\label{subsubsec:hyper}
The weights $\lambda_\text{freq}$ and $\lambda_\text{bin}$ were searched over the range $\{0, 0.01, 0.1, 1.0\}$ and $\{0, 0.001, 0.01, 0.1 \}$ in clustering, respectively. For the multi-layer setup, we conduct a sequential grid search in range $\{0, 0.0001, 0.001, 0.01, 0.1\}$ following the order $\lambda_\text{empty}$, $\lambda_\text{freq}$, and $\lambda_\text{bin}$. We provide the sequential grid search results of $\lambda_\text{freq}$ and $\lambda_\text{bin}$ on four characteristically distinct datasets, ColoredMNIST~\cite{arjovsky2019invariant}, FRGC~\cite{phillips2005frgc}, SVHN~\cite{netzer2011reading}, and GTSRB-8~\cite{stallkamp2012man}, to demonstrate the sensibility of our model to these hyperparameters in~\cref{tab:grid}. Our results indicate that $\lambda_\text{freq}$ is a critical hyperparameter for preventing cluster collapse. We observed that $\lambda_\text{bin}$ acts as a regularizer that improves the probabilities to be one-hot, but has a lower impact on overall performance compared to $\lambda_\text{freq}$. Although tuning regularization hyperparameters via ground truth labels allows us to establish a performance upper bound for the proposed architecture, we acknowledge that this protocol departs from a strictly unsupervised setting. We identify the development of robust, fully unsupervised model selection criteria as a significant remaining challenge for the field.
\begin{table}
\centering
\caption{\yellowcircle{} \textbf{Effect of $\lambda_{\text{freq}}$ (left) and $\lambda_{\text{bin}}$ (right)} on four datasets: ColoredMNIST~\cite{arjovsky2019invariant}, FRGC~\cite{phillips2005frgc}, SVHN~\cite{netzer2011reading}, and GTSRB-8~\cite{stallkamp2012man}.  (\orangecircle{}) Gumbel softmax, (\yellowcircle{}) $\lambda_{\text{bin}}=0$ (left), $\mathcal{L}^s_{\text{rec}}$.}
    \begin{subtable}[b]{.49\linewidth}
    \centering
    \resizebox{\textwidth}{!}
    {\begin{tabular}{@{}lcccc@{}}
    \toprule
    & \multicolumn{4}{c}{$\lambda_{\text{freq}}$} \\
    \cmidrule{2-5}
    Dataset & 1.0 & 0.1 & 0.01 & 0 \\
    \midrule
    ColoredMNIST & 82.7±2.2 & \textbf{95.9±0.1} & 86.4±2.6 & 53.2±4.3 \\
    \midrule
    FRGC & - & 36.1±0.8 & \textbf{44.8±0.8} & 39.5±1.3 \\
    \midrule
    SVHN & - & 32.8±0.7 & \textbf{35.3±0.4} & 33.9±0.5 \\
    \midrule
    GTSRB-8 & 53.0±1.15 & \textbf{53.2±1.2} & 50.5±0.7 & 50.0±0.2 \\
    \bottomrule
    \end{tabular}}
    \end{subtable}
    \hfill
    \begin{subtable}[b]{.458\linewidth}
    \centering
    \resizebox{\textwidth}{!}
    {\begin{tabular}{@{}l l c c c @{}}
    \toprule
    & & \multicolumn{3}{c}{$\lambda_{\text{bin}}$}\\
    \cmidrule{3-5}
    Dataset & \color{gray} $\lambda_{\text{freq}}$ & 0.01 & 0.001 & 0\\
    \midrule
    ColoredMNIST &  \color{gray} 0.1 & 56.6±9.3 & \textbf{96.0±0.1} & 95.9±0.1 \\
    \midrule
    FRGC & \color{gray} 0.01 & 42.5±1.0 & 41.7±0.6 & \textbf{44.8±0.8} \\
    \midrule
    SVHN & \color{gray} 0.01 & 36.1±0.5 & \textbf{37.6±0.3} & 35.3±0.4 \\
    \midrule
    GTSRB-8 & \color{gray} 0.1 & 49.7±0.3 & 50.0±0.8 & \textbf{53.2±1.2} \\
    \bottomrule
    \end{tabular}}
    \end{subtable}
    \label{tab:grid}
\end{table}


%% file: ref.bib
@inproceedings{bottou1994convergence,
  author = {Bottou, L\'{e}on and Bengio, Yoshua},
  title = {Convergence properties of the {K}-means algorithms},
  booktitle = {Advances in Neural Information Processing Systems},
  year = {1994}
}

@article{jiang2020generative,
  title={Generative neurosymbolic machines},
  author={Jiang, Jindong and Ahn, Sungjin},
  journal={Advances in Neural Information Processing Systems},
  year={2020}
}

@article{greff2020binding,
  title={On the binding problem in artificial neural networks},
  author={Greff, Klaus and Van Steenkiste, Sjoerd and Schmidhuber, J{\"u}rgen},
  journal={arXiv preprint arXiv:2012.05208},
  year={2020}
}

@inproceedings{stelzner2019faster,
  title={Faster attend-infer-repeat with tractable probabilistic models},
  author={Stelzner, Karl and Peharz, Robert and Kersting, Kristian},
  booktitle={Proceedings of the International Conference on Machine Learning},
  year={2019}
}

@inproceedings{singh2025glass,
  title={{GLASS}: Guided Latent Slot Diffusion for Object-Centric Learning},
  author={Singh, Krishnakant and Schaub-Meyer, Simone and Roth, Stefan},
  booktitle={Proceedings of the IEEE/CVF Conference on Computer Vision and Pattern Recognition},
  year={2025}
}

@article{wu2023slotdiffusion,
  title={{SlotDiffusion}: Object-centric generative modeling with diffusion models},
  author={Wu, Ziyi and Hu, Jingyu and Lu, Wuyue and Gilitschenski, Igor and Garg, Animesh},
  journal={Advances in Neural Information Processing Systems},
  year={2023}
}

@article{jiang2023object,
  title={Object-centric slot diffusion},
  author={Jiang, Jindong and Deng, Fei and Singh, Gautam and Ahn, Sungjin},
  journal={arXiv preprint arXiv:2303.10834},
  year={2023}
}

@inproceedings{macqueen1967some,
  title={Some methods for classification and analysis of multivariate observations},
  author={MacQueen, James},
  booktitle={Proceedings of the Fifth Berkeley Symposium on Mathematical Statistics and Probability, Volume 1: Statistics},
  organization={University of California press},
  year={1967}
}

@inproceedings{jang2017categorical,
    title={Categorical Reparameterization with {Gumbel}-Softmax},
    author={Eric Jang and Shixiang Gu and Ben Poole},
    booktitle={Proceedings of the International Conference on Learning Representations},
    year={2017}
}

@article{villa2024unsupervised,
  title={Unsupervised Object Discovery: A Comprehensive Survey and Unified Taxonomy},
  author={Villa-V{\'a}squez, Jos{\'e}-Fabian and Pedersoli, Marco},
  journal={arXiv preprint arXiv:2411.00868},
  year={2024}
}

@inproceedings{sauvalle2023unsupervised,
    title={Unsupervised multi-object segmentation using attention and soft-argmax},
    author={Sauvalle, Bruno and de La Fortelle, Arnaud},
    booktitle={Proceedings of the IEEE/CVF Winter Conference on Applications of Computer Vision},
    year={2023}
}

@inproceedings{karazija2021clevrtex,
    author={Laurynas Karazija and Iro Laina and Christian Rupprecht},
    booktitle={Advances in Neural Information Processing Systems Datasets and Benchmarks Track},
    title={{C}levr{T}ex: A Texture-Rich Benchmark for Unsupervised Multi-Object Segmentation},
    year={2021}
}

@inproceedings{maddison2017concrete,
    title={{The Concrete Distribution}: A Continuous Relaxation of Discrete Random Variables},
    author={Chris J. Maddison and Andriy Mnih and Yee Whye Teh},
    booktitle={Proceedings of the International Conference on Learning Representations},
    year={2017},
}

@article{arjovsky2019invariant,
  title={Invariant risk minimization},
  author={Arjovsky, Martin and Bottou, L{\'e}on and Gulrajani, Ishaan and Lopez-Paz, David},
  journal={arXiv preprint arXiv:1907.02893},
  year={2019}
}

@article{xiao2017fashion,
    title={Fashion-{MNIST}: a novel image dataset for benchmarking machine learning algorithms},
    author={Xiao, Han and Rasul, Kashif and Vollgraf, Roland},
    journal={arXiv preprint arXiv:1708.07747},
    year={2017}
}

@article{stallkamp2012man,
    title = {Man vs. computer: Benchmarking machine learning algorithms for traffic sign recognition},
    author = {J. Stallkamp and M. Schlipsing and J. Salmen and C. Igel},
    journal = {Neural Networks},
    year = {2012}
}

@article{netzer2011reading,
    title={Reading digits in natural images with unsupervised feature learning},
    journal={Advances in Neural Information Processing Systems Workshop on Deep Learning and Unsupervised Feature Learning},
    author={Netzer, Yuval and Wang, Tao and Coates, Adam and Bissacco, Alessandro and Wu, Bo and Ng, Andrew Y},
    year={2011}
}

@misc{lecun2010mnist,
    title={{MNIST} handwritten digit database},
    author={LeCun, Yann and Cortes, Corinna and Burges, Chris and others},
    publisher={Florham Park, NJ, USA},
    year={2010}
}

@article{hull1994usps,
    author={Hull, J.J.},
    journal={IEEE Transactions on Pattern Analysis and Machine Intelligence}, 
    title={A database for handwritten text recognition research}, 
    year={1994}
}

@inproceedings{phillips2005frgc,
    author={Phillips, P.J. and Flynn, P.J. and Scruggs, T. and Bowyer, K.W. and Jin Chang and Hoffman, K. and Marques, J. and Jaesik Min and Worek, W.},
    booktitle={Proceedings of the IEEE/CVF Conference on Computer Vision and Pattern Recognition}, 
    title={Overview of the face recognition grand challenge}, 
    year={2005}
}

@inproceedings{monnier2021unsupervised,
    title={Unsupervised Layered Image Decomposition into Object Prototypes},
    author={Monnier, Tom and Vincent, Elliot and Ponce, Jean and Aubry, Mathieu},
    booktitle={Proceedings of the IEEE/CVF International Conference on Computer Vision},
    year={2021}
}

@article{smirnov2021marionette,
    title={Mario{N}ette: Self-supervised sprite learning},
    author={Smirnov, Dmitriy and Gharbi, Michael and Fisher, Matthew and Guizilini, Vitor and Efros, Alexei and Solomon, Justin M},
    journal={Advances in Neural Information Processing Systems},
    year={2021}
}

@inproceedings{monnier2020dticlustering,
    title={Deep Transformation-Invariant Clustering},
    author={Monnier, Tom and Groueix, Thibault and Aubry, Mathieu},
    booktitle={Advances in Neural Information Processing Systems},
    year={2020}
}

@article{loiseau2023learnable,
    title={{Learnable Earth Parser}: Discovering 3D Prototypes in Aerial Scans},
    author={Loiseau, Romain and Vincent, Elliot and Aubry, Mathieu and Landrieu, Loic},
    journal={Proceedings of the IEEE/CVF Conference on Computer Vision and Pattern Recognition},
    year={2024}
}

@inproceedings{ronneberger2015u,
    title={{U-Net}: Convolutional networks for biomedical image segmentation},
    author={Ronneberger, Olaf and Fischer, Philipp and Brox, Thomas},
    booktitle={International Conference on Medical Image Computing and Computer-Assisted Intervention},
    pages={234--241},
    year={2015},
    organization={Springer}
}

@inproceedings{siglidis2024ltr,
    title={{The Learnable Typewriter}: A Generative Approach to Text Analysis},
    author={Ioannis Siglidis and Nicolas Gonthier and Julien Gaubil and Tom Monnier and Mathieu Aubry},
    booktitle={Proceedings of the International Conference on Document Analysis and Recognition},
    year={2024},
}

@article{jaderberg2015spatial,
    title={Spatial transformer networks},
    author={Jaderberg, Max and Simonyan, Karen and Zisserman, Andrew and others},
    journal={Advances in Neural Information Processing Systems},
    year={2015}
}

@inproceedings{frey99em,
    title={Estimating mixture models of images and inferring spatial transformations using the {EM} algorithm}, 
    author={Frey, B.J. and Jojic, N.},
    booktitle={Proceedings of the IEEE/CVF Conference on Computer Vision and Pattern Recognition}, 
    year={1999}
}

@inproceedings{frey01fltic,
    title = {Fast, Large-Scale Transformation-Invariant Clustering},
    author = {Frey, Brendan J and Jojic, Nebojsa},
    booktitle = {Advances in Neural Information Processing Systems},
    publisher = {MIT Press},
    year = {2001}
}

@article{frey03tic,
    title={Transformation-invariant clustering using the {EM} algorithm}, 
    author={Frey, B.J. and Jojic, N.},
    journal={IEEE Transactions on Pattern Analysis and Machine Intelligence}, 
    year={2003}
}

@article{lm06data,
    author={Learned-Miller, E.G.},
    journal={IEEE Transactions on Pattern Analysis and Machine Intelligence}, 
    title={Data driven image models through continuous joint alignment}, 
    year={2006}
}

@inproceedings{huang07uja,
    author={Huang, Gary B. and Jain, Vidit and Learned-Miller, Erik},
    booktitle={Proceedings of the IEEE/CVF International Conference on Computer Vision}, 
    title={Unsupervised Joint Alignment of Complex Images}, 
    year={2007}
}

@inproceedings{miller00shared,
    author={Miller, E.G. and Matsakis, N.E. and Viola, P.A.},
    booktitle={Proceedings of the IEEE/CVF Conference on Computer Vision and Pattern Recognition}, 
    title={Learning from one example through shared densities on transforms}, 
    year={2000}
}

@article{annunziata2019jointly,
    title={Jointly Aligning Millions of Images With Deep Penalised Reconstruction Congealing},
    author={Roberto Annunziata and Christos Sagonas and Jacques Cal{\`i}},
    journal={Proceedings of the IEEE/CVF International Conference on Computer Vision},
    year={2019}
}

@inproceedings{cox2008lsc,
    title           = "Least squares congealing for unsupervised alignment of images",
    booktitle       = "Proceedings of the IEEE/CVF Conference on Computer Vision and Pattern Recognition",
    author          = "Cox, Mark and Sridharan, Sridha and Lucey, Simon and Cohn, Jeffrey",
    year            =  2008,
}

@inproceedings{cox2009lsc,
    author = {Cox, M. and Sridharan, S. and Lucey, S. and Cohn, J.},
    title = {Least-Squares Congealing for Large Numbers of Images},
    booktitle = {Proceedings of the IEEE/CVF International Conference on Computer Vision},
    year = {2009}
}

@inproceedings{mattar2012unsupervised,
    author = {Mattar, Marwan A. and Hanson, Allen R. and Learned-Miller, Erik G.},
    title = {Unsupervised joint alignment and clustering using Bayesian nonparametrics},
    booktitle = {Conference on Uncertainty in Artificial Intelligence},
    year = {2012},
}

@inproceedings{liu2009simal,
    title={Simultaneous alignment and clustering for an image ensemble}, 
    author={Liu, Xiaoming and Tong, Yan and Wheeler, Frederick W.},
    booktitle={Proceedings of the IEEE/CVF International Conference on Computer Vision}, 
    year={2009}
}

@article{zhou2024comprehensive,
    author = {Zhou, Sheng and Xu, Hongjia and Zheng, Zhuonan and Chen, Jiawei and Li, Zhao and Bu, Jiajun and Wu, Jia and Wang, Xin and Zhu, Wenwu and Ester, Martin},
    title = {{A Comprehensive Survey on Deep Clustering}: Taxonomy, Challenges, and Future Directions},
    journal = {ACM Computing Surveys},
    year = {2024},
}

@article{ren2024deep,
    author={Ren, Yazhou and Pu, Jingyu and Yang, Zhimeng and Xu, Jie and Li, Guofeng and Pu, Xiaorong and Yu, Philip S. and He, Lifang},
    journal={IEEE Transactions on Neural Networks and Learning Systems}, 
    title={{Deep Clustering}: A Comprehensive Survey}, 
    year={2024}
}

@article{wei2024odc,
    author = {Xiuxi Wei and Zhihui Zhang and Huajuan Huang and Yongquan Zhou},
    title = {An overview on deep clustering},
    journal = {Neurocomputing},
    year = {2024}
}

@inproceedings{haeusser2019adc,
    author="Haeusser, Philip and Plapp, Johannes and Golkov, Vladimir and Aljalbout, Elie and Cremers, Daniel",
    editor="Brox, Thomas and Bruhn, Andr{\'e}s and Fritz, Mario",
    title="{Associative Deep Clustering}: Training a Classification Network with No Labels",
    booktitle="Pattern Recognition",
    year="2019"
}

@inproceedings{caron2018deep,
    title={Deep Clustering for Unsupervised Learning of Visual Features},
    author={Mathilde Caron and Piotr Bojanowski and Armand Joulin and Matthijs Douze},
    booktitle={Proceedings of the IEEE/CVF European Conference on Computer Vision},
    year={2018},
}

@inproceedings{caron2020unsupervised,
    title={Unsupervised Learning of Visual Features by Contrasting Cluster Assignments},
    author={Caron, Mathilde and Misra, Ishan and Mairal, Julien and Goyal, Priya and Bojanowski, Piotr and Joulin, Armand},
    booktitle={Advances in Neural Information Processing Systems},
    year={2020}
}

@inproceedings{li2021prototypical,
    title={Prototypical Contrastive Learning of Unsupervised Representations},
    author={Junnan Li and Pan Zhou and Caiming Xiong and Steven Hoi},
    booktitle={Proceedings of the International Conference on Learning Representations},
    year={2021},
}

@inproceedings{liang2023clusterfomer,
    title={{ClusterFormer}: Clustering As A Universal Visual Learner},
    author={James Chenhao Liang and Yiming Cui and Qifan Wang and Tong Geng and Wenguan Wang and Dongfang Liu},
    booktitle={Advances in Neural Information Processing Systems},
    year={2023}
}

@article{yang2016joint,
    title={Joint Unsupervised Learning of Deep Representations and Image Clusters},
    author={Jianwei Yang and Devi Parikh and Dhruv Batra},
    journal={Proceedings of the IEEE/CVF Conference on Computer Vision and Pattern Recognition},
    year={2016}
}

@inproceedings{chang2017deep,
    author={Chang, Jianlong and Wang, Lingfeng and Meng, Gaofeng and Xiang, Shiming and Pan, Chunhong},
    booktitle={Proceedings of the IEEE/CVF International Conference on Computer Vision}, 
    title={Deep Adaptive Image Clustering}, 
    year={2017}
}

@article{mrabah2019adversarial,
    title={{Adversarial Deep Embedded Clustering}: On a Better Trade-off Between Feature Randomness and Feature Drift},
    author={Nairouz Mrabah and Mohamed Bouguessa and Riadh Ksantini},
    journal={IEEE Transactions on Knowledge and Data Engineering},
    year={2019}
}

@inproceedings{xie2016unsupervised,
    title={Unsupervised deep embedding for clustering analysis},
    author={Xie, Junyuan and Girshick, Ross and Farhadi, Ali},
    booktitle={Proceedings of the International Conference on Machine Learning},
    year={2016}
}

@inproceedings{ji2019invariant,
    title={Invariant information clustering for unsupervised image classification and segmentation},
    author={Ji, Xu and Henriques, Jo{\~a}o F and Vedaldi, Andrea},
    booktitle={Proceedings of the IEEE/CVF International Conference on Computer Vision},
    year={2019}
}

@inproceedings{hu2017learning,
    title={Learning discrete representations via information maximizing self-augmented training},
    author={Hu, Weihua and Miyato, Takeru and Tokui, Seiya and Matsumoto, Eiichi and Sugiyama, Masashi},
    booktitle={Proceedings of the International Conference on Machine Learning},
    year={2017}
}

@inproceedings{jiang2016variational,
    title={{Variational Deep Embedding}: An Unsupervised and Generative Approach to Clustering},
    author={Zhuxi Jiang and Yin Zheng and Huachun Tan and Bangsheng Tang and Hanning Zhou},
    booktitle={International Joint Conference on Artificial Intelligence},
    year={2016}
}

@inproceedings{mukherjee2018cluster,
    title={Cluster{GAN}: Latent Space Clustering in Generative Adversarial Networks},
    author={Sudipto Mukherjee and Himanshu Asnani and Eugene Lin and Sreeram Kannan},
    booktitle={Proceedings of the AAAI Conference on Artificial Intelligence},
    year={2018}
}

@article{dizaji2017deep,
    title={Deep Clustering via Joint Convolutional Autoencoder Embedding and Relative Entropy Minimization},
    author={Kamran Ghasedi Dizaji and Amirhossein Herandi and Cheng Deng and Weidong (Tom) Cai and Heng Huang},
    journal={Proceedings of the IEEE/CVF International Conference on Computer Vision},
    year={2017}
}

@article{niu2022spice,
    author={Niu, Chuang and Shan, Hongming and Wang, Ge},
    journal={IEEE Transactions on Image Processing}, 
    title={{SPICE}: Semantic Pseudo-Labeling for Image Clustering}, 
    year={2022}
}

@inproceedings{van2020scan,
    title={{SCAN}: Learning to classify images without labels},
    author={Van Gansbeke, Wouter and Vandenhende, Simon and Georgoulis, Stamatios and Proesmans, Marc and Van Gool, Luc},
    booktitle={Proceedings of the IEEE/CVF European Conference on Computer Vision},
    year={2020}
}

@inproceedings{kosiorek2019stacked,
	title = {Stacked Capsule Autoencoders},
	author = {Kosiorek, Adam R. and Sabour, Sara and Teh, Yee Whye and Hinton, Geoffrey E.},
	booktitle = {Advances in Neural Information Processing Systems},
	year = {2019},
}

@inproceedings{greff2019multiobject,
	title = {Multi-Object Representation Learning with Iterative Variational Inference},
	author = {Greff, Klaus and Kaufman, Rapha{\" e}l Lopez and Kabra, Rishabh and Watters, Nick and Burgess, Chris and Zoran, Daniel and Matthey, Loic and Botvinick, Matthew M. and Lerchner, Alexander},
	booktitle = {Proceedings of the International {Conference} on {Machine} {Learning}},
	year = {2019},
}

@inproceedings{crawford2019spatially,  
    title={Spatial Invariant Unsupervised Object Detection with Convolutional Neural Networks},  
    author={Crawford, Eric and Pineau, Joelle},  
    booktitle={Proceedings of the AAAI Conference on Artificial Intelligence},  
    year={2019}
}

@inproceedings{kosiorek2018sequential,
	title = {{Sequential Attend, Infer, Repeat}: Generative Modelling of Moving Objects},
	author = {Kosiorek, Adam R. and Kim, Hyunjik and Teh, Yee Whye and Posner, Ingmar},
	booktitle = {Advances in Neural Information Processing Systems},
	year = {2018},
}

@article{xiang2021dpr,
  title={{DPR-CAE}: capsule autoencoder with dynamic part representation for image parsing},
  author={Xiang, Canqun and Wang, Zhennan and Zou, Wenbin and Xu, Chen},
  journal={arXiv preprint arXiv:2104.14735},
  year={2021}
}

@conference{kugelgen2020towards,
    title = {Towards causal generative scene models via competition of experts},
    author = {von K{\"u}gelgen, J. and Ustyuzhaninov, I. and Gehler, P. and Bethge, M. and Sch{\"o}lkopf, B.},
    booktitle = {International Conference on Learning Representations Workshop on Causal Learning for Decision Making},
    year = {2020},
}

@inproceedings{yang2019deep,
  title={Deep spectral clustering using dual autoencoder network},
  author={Yang, Xu and Deng, Cheng and Zheng, Feng and Yan, Junchi and Liu, Wei},
  booktitle={Proceedings of the IEEE/CVF Conference on Computer Vision and Pattern Recognition},
  year={2019}
}

@inproceedings{
    shaham2018spectralnet,
    title={{SpectralNet}: Spectral Clustering using Deep Neural Networks},
    author={Uri Shaham and Kelly Stanton and Henry Li and Ronen Basri and Boaz Nadler and Yuval Kluger},
    booktitle={Proceedings of the International Conference on Learning Representations},
    year={2018},
}

@inproceedings{greff2017neural,
	title = {Neural Expectation Maximization},
	author = {Greff, Klaus and Steenkiste, Sjoerd van and Schmidhuber, J{\" u}rgen},
	booktitle = {Proceedings of the International {Conference} on {Learning} {Representations}},
	year = {2017},
}

@inproceedings{johnson2017clevr,
    title={{CLEVR}: A diagnostic dataset for compositional language and elementary visual reasoning},
    author={Johnson, Justin and Hariharan, Bharath and Van Der Maaten, Laurens and Fei-Fei, Li and Lawrence Zitnick, C and Girshick, Ross},
    booktitle={Proceedings of the IEEE/CVF Conference on Computer Vision and Pattern Recognition},
    year={2017}
}

@article{kuhn1955hungarian,
    author = {Kuhn, H. W.},
    title = {The {H}ungarian method for the assignment problem},
    journal = {Naval Research Logistics Quarterly},
    year = {1955}
}

@misc{kabra2019multi,
    title={Multi-Object Datasets},
    author={Kabra, Rishabh and Burgess, Chris and Matthey, Loic and
          Kaufman, Raphael Lopez and Greff, Klaus and Reynolds, Malcolm and
          Lerchner, Alexander},
    howpublished={https://github.com/deepmind/multi-object-datasets/},
    year={2019}
}

@inproceedings{kakogeorgiou2024spot,
	title = {{SPOT}: Self-Training with Patch-Order Permutation for Object-Centric Learning with Autoregressive Transformers},
	author = {Kakogeorgiou, Ioannis and Gidaris, Spyros and Karantzalos, Konstantinos and Komodakis, Nikos},
	booktitle = {Proceedings of the IEEE/CVF Conference on Computer Vision and Pattern Recognition},
	year = {2024},
}

@inproceedings{lin2020space,
	title = {{SPACE}: Unsupervised Object-Oriented Scene Representation via Spatial Attention and Decomposition},
	author = {Lin, Zhixuan and Wu, Yi-Fu and Peri, Skand Vishwanath and Sun, Weihao and Singh, Gautam and Deng, Fei and Jiang, Jindong and Ahn, Sungjin},
	booktitle = {International {Conference} on {Learning} {Representations}},
	year = {2020},
}

@article{engelcke2021genesis,
  title={{GENESIS-v2}: Inferring unordered object representations without iterative refinement},
  author={Engelcke, Martin and Parker Jones, Oiwi and Posner, Ingmar},
  journal={Advances in Neural Information Processing Systems},
  year={2021}
}

@inproceedings{emami2021efficient,
  title={Efficient iterative amortized inference for learning symmetric and disentangled multi-object representations},
  author={Emami, Patrick and He, Pan and Ranka, Sanjay and Rangarajan, Anand},
  booktitle={Proceedings of the International Conference on Machine Learning},
  year={2021}
}

@inproceedings{engelcke2020genesis,
	title = {{GENESIS}: Generative Scene Inference and Sampling with Object-Centric Latent Representations},
	author = {Engelcke, Martin and Kosiorek, Adam R. and Jones, Oiwi Parker and Posner, Ingmar},
	booktitle = {Proceedings of the International {Conference} on {Learning} {Representations}},
	year = {2020},
}

@inproceedings{visser2019stampnet,
	title = {{StampNet}: Unsupervised Multi-Class Object Discovery},
	author = {Visser, Joost and Corbetta, Alessandro and Menkovski, Vlado and Toschi, Federico},
	booktitle = {Proceedings of the IEEE {International} {Conference} on {Image} {Processing}},
	year = {2019},
}

@article{burgess2019monet,
  title={{MoNet}: Unsupervised scene decomposition and representation},
  author={Burgess, Christopher P and Matthey, Loic and Watters, Nicholas and Kabra, Rishabh and Higgins, Irina and Botvinick, Matt and Lerchner, Alexander},
  journal={arXiv preprint arXiv:1901.11390},
  year={2019}
}

@inproceedings{greff2016tagger,
	title = {Tagger: Deep Unsupervised Perceptual Grouping},
	author = {Greff, Klaus and Rasmus, Antti and Berglund, Mathias and Hao, Tele Hotloo and Valpola, Harri and Schmidhuber, J{\" u}rgen},
	booktitle = {Advances in Neural Information Processing Systems},
	year = {2016},
}

@inproceedings{eslami2016attend,
	title = {{Attend, Infer, Repeat}: Fast Scene Understanding with Generative Models},
	author = {Eslami, S. M. Ali and Heess, Nicolas and Weber, Theophane and Tassa, Yuval and Szepesvari, David and Kavukcuoglu, Koray and Hinton, Geoffrey E.},
	booktitle = {Advances in Neural Information Processing Systems},
	year = {2016},
}

@article{greff2015binding,
  title={Binding via reconstruction clustering},
  author={Greff, Klaus and Srivastava, Rupesh Kumar and Schmidhuber, J{\"u}rgen},
  journal={arXiv preprint arXiv:1511.06418},
  year={2015}
}

@inproceedings{vincent2008extracting,
    title={Extracting and composing robust features with denoising autoencoders},
    author={Pascal Vincent and H. Larochelle and Yoshua Bengio and Pierre-Antoine Manzagol},
    booktitle={Proceedings of the International Conference on Machine Learning},
    year={2008},
}

@inproceedings{seitzer2023bridging,
	title = {Bridging the Gap to Real-World Object-Centric Learning},
	author = {Seitzer, Maximilian and Horn, Max and Zadaianchuk, Andrii and Zietlow, Dominik and Xiao, Tianjun and Simon-Gabriel, Carl-Johann and He, Tong and Zhang, Zheng and Sch{\" o}lkopf, Bernhard and Brox, Thomas and Locatello, Francesco},
	booktitle = {Proceedings of the International {Conference} on {Learning} {Representations}},
	year = {2023},
}

@inproceedings{locatello2020objectcentric,
	title = {Object-Centric Learning with Slot Attention},
	author = {Locatello, Francesco and Weissenborn, Dirk and Unterthiner, Thomas and Mahendran, Aravindh and Heigold, Georg and Uszkoreit, Jakob and Dosovitskiy, Alexey and Kipf, Thomas},
	booktitle = {Advances in Neural Information Processing Systems},
	year = {2020},
}

@inproceedings{caron2021emerging,
    author    = {Caron, Mathilde and Touvron, Hugo and Misra, Ishan and J\'egou, Herv\'e and Mairal, Julien and Bojanowski, Piotr and Joulin, Armand},
    title     = {Emerging Properties in Self-Supervised Vision Transformers},
    booktitle = {Proceedings of the IEEE/CVF International Conference on Computer Vision},
    year      = {2021},
}

@inproceedings{singh2022illiterate,
    title={Illiterate {DALL-E} Learns to Compose},
    author={Gautam Singh and Fei Deng and Sungjin Ahn},
    booktitle={Proceedings of the International Conference on Learning Representations},
    year={2022},
}

@misc{affnist,
	author = {Tijmen Tieleman},
	title = {aff{N}{I}{S}{T} --- cs.toronto.edu},
	howpublished = {\url{https://www.cs.toronto.edu/~tijmen/affNIST/}},
	year = {2013},
	note = {[Accessed 04-11-2025]},
}
